\documentclass[runningheads]{llncs}

\usepackage{eccv}

\usepackage{eccvabbrv}

\usepackage{graphicx}
\usepackage{booktabs}
\usepackage{algorithm}
\usepackage{algorithmic}
\usepackage{multirow}
\usepackage{tabularx}
\usepackage{wrapfig}
\usepackage[accsupp]{axessibility}  

\usepackage{hyperref}

\usepackage{orcidlink}

\begin{document}
\begin{sloppypar}
\title{Training-free Subject-Enhanced Attention Guidance for Compositional
Text-to-image Generation} 

\titlerunning{Abbreviated paper title}

\author{Shengyuan Liu\inst{1,2} \and
Bo Wang \inst{3} \and
Ye Ma \inst{3} \and Te Yang \inst{1,2} \and Xipeng Cao \inst{3} \and Quan Chen \inst{3} \and Han Li \inst{3} \and Di Dong\inst{1,2} \and Peng Jiang \inst{3}}

\authorrunning{Shengyuan Liu, Bo Wang et al.}

\institute{University of Chinese Academy of Sciences, Beijing, China \and Institute of Automation, Chinese Academy of Sciences, Beijing, China \and
Kuaishou Technology, Beijing, China
}

\maketitle

\begin{abstract}
Existing subject-driven text-to-image generation models suffer from tedious fine-tuning steps and struggle to maintain both text-image alignment and subject fidelity. For generating compositional subjects, it often encounters problems such as object missing and attribute mixing, where some subjects in the input prompt are not generated or their attributes are incorrectly combined. To address these limitations, we propose a subject-driven generation framework and introduce training-free guidance to intervene in the generative process during inference time. This approach strengthens the attention map, allowing for precise attribute binding and feature injection for each subject. Notably, our method exhibits exceptional zero-shot generation ability, especially in the challenging task of compositional generation. Furthermore, we propose a novel metric GroundingScore to evaluate subject alignment thoroughly. The obtained quantitative results serve as compelling evidence showcasing the effectiveness of our proposed method. The code will be released soon.
\end{abstract}

\section{Introduction}
\label{sec:intro}

Text-to-image generation models have made significant advancements and enabled the creation of diverse and high-quality images based on textual prompts. Several robust diffusion-based models, such as GLIDE \cite{glide}, Imagen \cite{imagen}, Stable Diffusion \cite{stablediffusion}, and DALL-E3 \cite{DALLE2}, exhibit the capability to comprehend semantic information conveyed through textual prompts, facilitating the generation of superior-quality images that align with the provided descriptions. 

In the subject-driven generation task, the focus is not solely on generating diverse images but rather on customizing new images of a specific subject by using one or multiple images of the subject along with textual prompts. This customization includes variations in poses, backgrounds, object locations, dressing, lighting, and styles while maintaining the same identity.

\begin{figure}[t]
  \centering
   \includegraphics[width=0.85\linewidth]{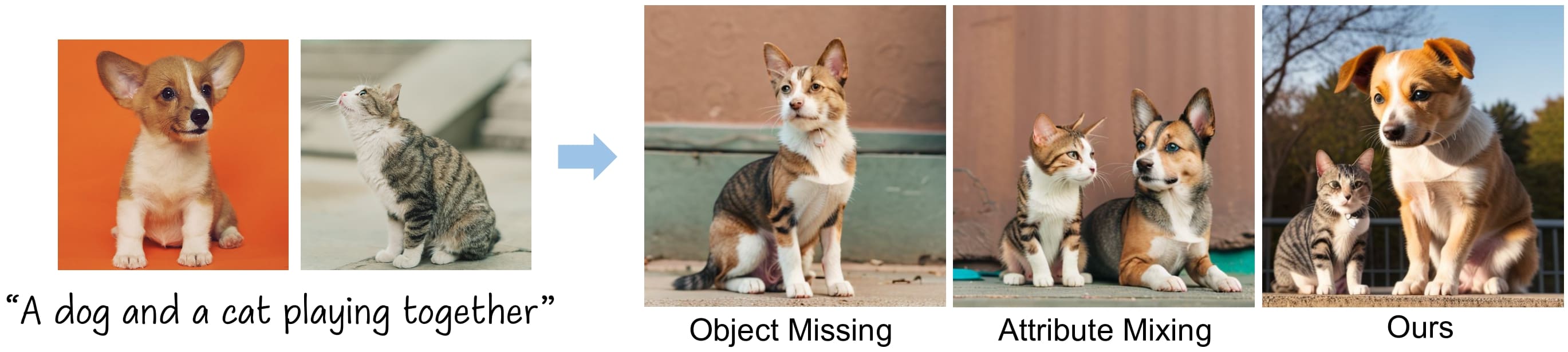}
   \caption{\textbf{Compositional subject-driven generation.} Existing text-to-image models frequently encounter difficulties in multi-subject generation tasks due to issues such as "object missing" and "attribute mixing", where some subjects in the prompt are not generated or their attributes are incorrectly combined. To address these challenges, our proposed SE-Guidance method effectively preserves the attributes of each subject, ensuring accurate and reliable generation outcomes.}
   \label{fig:example}
\end{figure}

Currently, there are two main categories of methods for this task. The common approaches involve inverting subject visuals into text embedding space, typically using a unique identifier or a placeholder \cite{Dreambooth, TextualInversion, customdiffusion}. These embeddings are then composed into natural language prompts to create different renditions of the subject. However, a known inefficiency of this way is the requirement of hundreds or thousands of tedious fine-tuning steps for each new subject, which hinders its scalability to a wide range of subjects. Other approaches \cite{Blipdiffusion, instantbooth, ELITE} involve extracting the representation of the image subject and concatenating it with text features without fine-tuning. However, this method suffers from the existing gap between the textual and image modalities, which can easily disrupt the original semantic information and lead to uncontrolled results. 

Compositional generation, as a subtask of subject-driven generation, presents greater challenges to model performance. Existing approaches often exhibit "object missing" where one or more subjects in the prompt are not generated, and "attribute mixing" wherein the attributes of the subjects are incorrectly combined or mixed, as exemplified in \cref{fig:example}. We attribute this issue to the difficulty in establishing a strong correlation between image features and the corresponding textual semantics associated with each subject, as well as the precise injection of subject features into generated images.

To enable controllable and high-fidelity generation, we propose to utilize an image prompt adapter with a subject-representation encoder to avoid the shortcomings of the previous methods. The adapter injects subject attributes into the diffusion model without modifying the original text-to-image models. This model structure is generalizable for both single and compositional subject-driven generation, with the ability to perform zero-shot or minimal fine-tuning steps. Furthermore, it supports the input of multiple images depicting the same subject.

Furthermore, we present an intuitive and effective method named \textbf{S}ubject-\textbf{E}nhanced Attention Guidance (SE-Guidance) to improve the subject-driven text-to-image generation task, especially for compositional generation. This method leverages cross-attention maps to bind subject image features with their corresponding semantic information while disentangling subject pairs from each other. Importantly, our approach is applied on the fly during inference time and requires no additional training or fine-tuning.

During the evaluations, we observed existing evaluation metrics for subject image alignment are susceptible to interference from background information. When the generated image overfits the original image, the evaluation metric often gives erroneously high scores, failing to accurately represent subject fidelity, especially in multi-subject generation tasks. Inspired by the grounding task, we introduce a novel metric called GroundingScore to provide a more precise assessment and demonstrate the effectiveness of our methods on the subject-driven generation dataset CustomConcept101 \cite{customdiffusion}.

To sum up, our contributions are as follows:
\begin{itemize}
    \item We propose a zero-shot framework for subject-driven generation that effectively integrates image information using a disentangled cross-attention strategy. This framework achieves state-of-the-art results, particularly in compositional generation scenarios.
    \item We introduce a subject-enhanced attention guidance method that achieves precise attribute binding and feature injection for each subject. Our training-free method significantly reduces the subject-driven generation process time to within 15 seconds.
    \item We conduct experiments to validate the effectiveness of our model and propose new evaluation metrics for a more comprehensive assessment of the compositional text-to-image generation task.
\end{itemize}

\section{Related Works}
\label{sec:related}

\textbf{Text-to-image generation.} Early studies focused on text-to-image generation within the context of Generative Adversarial Networks \cite{GAN, gan4, biggan, Stylegan, qiao2019learn}. However, they are limited to generating well-aligned and highly structured objects, such as face completion \cite{dcgan,pggan,amtgan}. More recently, remarkable advancements have been made using large-scale auto-regressive models \cite{vqvae, vqvae2, tamingtransformer,attngan,Parti, MakeAScene} and diffusion models \cite{glide, imagen, stablediffusion, DALLE2, Uvit,vqdiff, RAPHAEL, ediffi}. Among them, GLIDE \cite{glide} employs a cascaded diffusion architecture with classifier-free guidance and demonstrates the synthesis of high-quality images through large-scale training. Stable Diffusion (SD) \cite{stablediffusion} trains a single diffusion model over a discretized latent space, allowing for the generation of high-resolution images with multiple conditions. After SD, many variations of SD models \cite{re-imagen, kandinsky, sdxl, Seecoder}and adapters \cite{controlnet, t2i-adapter, ip-adapter,uni-controlnet} have been proposed to enhance image quality and image-text alignment. However, these models face challenges in consistently preserving the identity of a subject across synthesized images.

\textbf{Subject-driven text-to-image generation.} Given a few images of a specific subject, existing subject-driven generation models \cite{Dreambooth, TextualInversion, disenbooth, customdiffusion, Blipdiffusion, SuTI} often employ a fine-tuning process using a pre-trained text-to-image diffusion model with a special token as a placeholder. For example, DreamBooth \cite{Dreambooth} fine-tunes the diffusion model, while Textual Inversion \cite{TextualInversion} fine-tunes the prompt embedding. This approach is time-consuming as it needs computational costs for every given subject. However, this approach is computationally expensive as it requires fine-tuning for every given subject. Some studies propose training additional modules to enable zero-shot or few-shot subject-driven generation, without the need for a fine-tuning step. These methods include the integration of a modified subject encoder \cite{Blipdiffusion, ELITE, tamingencoder} or an adapter \cite{instantbooth, designing} into the diffusion model. Despite these advancements, the existing methods tend to learn the subject's attributes in a tangled manner, potentially disrupting the original semantic information and altering the subject's identity in the generated image.

\textbf{Compositional subjects generation.} Existing subject-driven text-to-image models \cite{composablediffusion, Blipdiffusion, cosposition3, prompt2prompt} often exhibit object missing and attribute mixing issues. To address these issues, some works \cite{attendandexcite, layoutguidance, Cones, Cones2} incorporate additional guidance for compositional text-to-image generation. StructureDiffusion \cite{structurediffusion} utilizes structured representations of language inputs to enhance the CLIP \cite{CLIP} text embeddings for every subject. ZestGuide \cite{zestguide} leverages semantic maps as additional supervision to align implicit segmentation maps extracted from cross-attention layers. Our idea is to adaptively extract and enhance the potential spatial position information of the subjects from attention maps for accurate feature injection, without requiring an additional mask. When incorporating multi-subject images as conditional inputs, CustomDiffusion \cite{customdiffusion} fine-tunes the cross-attention layers and the prompts embeddings together for compositional generation. SVDiff \cite{svdiff} proposes a compact parameter space, spectral shift, for diffusion model fine-tuning. However, these approaches are inefficient when scaling to a wide range of subjects due to the fine-tuning step. Additionally, the challenge of generating subjects with identity preservation and language-image alignment remains unresolved in this task.

\section{Methods}
\label{sec:methods}

\begin{figure*}[t]
  \centering
   \includegraphics[width=\linewidth]{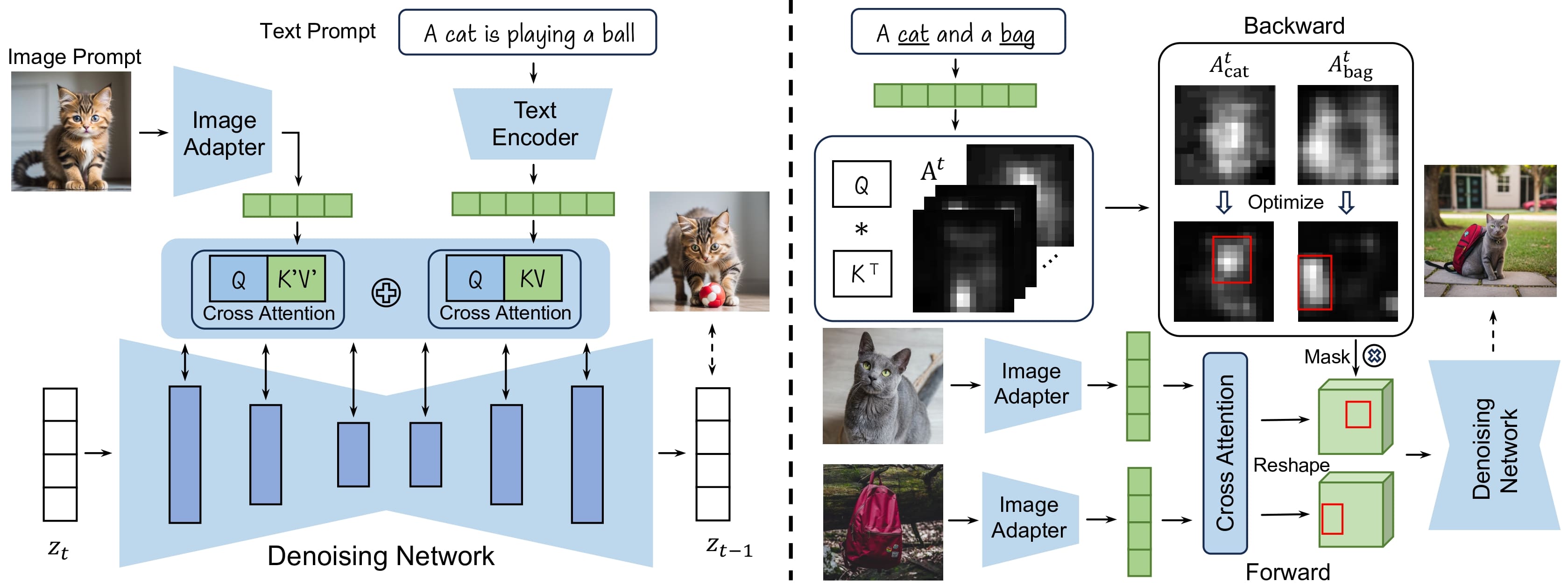}
   \caption{\textbf{Overall pipeline.} The left side is our proposed subject-driven generation framework that effectively integrates image information using a disentangled cross-attention strategy. On the right side, we illustrate the subject-enhanced attention guidance strategy, mainly including representation injection in the forward process and attention enhancement in the backward process.}
   \label{fig:pipeline}
\end{figure*}

This section introduces the preliminary concepts and presents the training-free subject-enhanced attention guidance strategy. An overview of our proposed SE-Guidance is shown in \cref{fig:pipeline}.
\subsection{Preliminaries}

\textbf{Stable diffusion models.} Diffusion models are probabilistic generative models that are trained to learn a data distribution by the gradual denoising of a variable sampled from a Gaussian distribution. We apply our method to the state-of-the-art Stable Diffusion (SD) \cite{stablediffusion} model.  Instead of operating in the image space, the denoising process of SD is conducted in the latent space by an auto-encoder. Specifically, an encoder $\mathcal{E}(\cdot)$ transforms the input image $x$ to latent variable $z = \mathcal{E}(x)$. During the denoising process, with a randomly sampled noise $\epsilon \sim \mathcal{N}(0, 1)$ and the time step $t$, we can get a noisy latent $z_t = \alpha_{t}z + \sigma_{t} \epsilon$, where $\alpha_t$ and $\sigma_t$ are the coefficients that control the noise schedule. Then the conditional diffusion model $\epsilon_\theta$ is optimized with the following objective: 

\begin{equation}
\mathcal{L} = \mathbb{E}_{z, c, \epsilon \sim \mathcal{N}(0,1), t}\left[\left\|\epsilon-\epsilon_\theta\left(z_t,  c, t\right)\right\|_2^2\right]
  \label{eq:diff}
\end{equation}
where $c$ represents the additional conditions, including text and image prompts. Generally, samplers such as DDPM \cite{DDPM}, DDIM \cite{DDIM}, PNDM \cite{PNDM} and DPM-Solver \cite{DPMSolver}, are adopted in the inference stage.

\textbf{Conditioned cross-attention in SD.} Cross-attention block modifies the latent features of the network according to the condition features, i.e., text features in the case of text-to-image diffusion models. Given text features $c \in \mathbb{R}^{l \times d}$ and latent image features $f\in \mathbb{R}^{(H \times W) \times C}$, a single-head cross-attention is calculated as:
\begin{equation}
Attn(Q, K, V)=Softmax\left(\frac{Q K^T}{\sqrt{d^{\prime}}}\right) V
\label{eq:attn}
\end{equation}
where $Q=W_q f, K=W_k c, V=W_v c$, and $d^{\prime}$ is the output dimension of key and query features. The latent feature is then updated with the attention block output. 

\subsection{Multi-modal Prompts Adapter}
Inspired by IP-Adapter \cite{ip-adapter}, we apply a multi-modal prompts adapter as shown in \cref{fig:pipeline}, which consists of two parts: an image encoder to extract subject representations and a cross-attention module to embed image features into the pre-trained SD model.

\textbf{Image encoder.} Given a few images of the target subjects, we use an image encoder to extract image features and inject them into the diffusion model. Following most of the methods, we use CLIP \cite{CLIP} as an image encoder, which is trained by contrastive learning on a large dataset of image-text pairs. Other methods such as a generic subject encoder in BLIP-Diffusion \cite{Blipdiffusion} is also compatible with the framework. Moreover, a trainable projection layer is employed to effectively align with the dimension of the text features. 

\textbf{Multi-modal cross-attention.} To inject subject representation while in the meantime largely inheriting the generation capabilities of the underlying diffusion model, we follow IP-Adapter and employ a new cross-attention layer with trainable parameters $W^{'}_k$ and $W^{'}_v$. The output of the attention block is defined as follows:
\begin{equation}
z = z^{T} + \lambda z^{I}
\label{eq:3}
\end{equation}
where $z^{T} = Attn(Q, K, V)$ and $z^{I} = Attn(Q,K',V')$ represent the text and image cross-attention outputs. $\lambda$ is a weight factor to control the strength of image feature injection.

IP-Adapter is trained using text-to-image pairs, posing a challenge in generating compositional images while ensuring subject consistency. To extend this framework for multi-subject generation tasks, where obtaining high-quality datasets with compositional subjects and related texts is difficult, we propose a training-free strategy to address this issue based on attention modules.

\subsection{Training-Free Subject-Enhanced Attention Guidance}
As shown on the right side of \cref{fig:pipeline}, our approach relies on a straightforward and intuitive concept: for a subject to appear in the synthesized image, it should have a significant impact on a particular patch in the image. Furthermore, the feature map associated with the subject image should also focus on that specific region. In this section, we introduce each step of our method.

\textbf{Subject attention-maps extraction.} First, given a text prompt including the subject tokens, we define the set of all subject tokens $S = \{ s_{1}, \cdots, s_{i}\}$. Our objective is to extract a spatial attention map for each token $s_i$, indicating the semantic intensity on each image patch. At each denoising step $t$, we perform a forward pass through the pre-trained conditional UNet with the noised latent $z_t$. For different downsampling resolutions, we consider averaging all corresponding attention maps in the UNet layers and heads. $A^{t} \in \mathbb{R}^{P^2 \times N}$ represents the attention maps of the text prompt with $N$ tokens in the denoising step $t$, where $P$ is the patch size in the transformer block. 


Then, we obtain the subjects' attention maps $\{ A^{t}_{s}\}$ from $A^{t}$ and apply a Gaussian smoothing filter, as the original maps may not fully capture the presence of objects in the generated image. To address this limitation, we conduct the smoothing operation to focus attention within specific regions, enhancing the generation of complete objects by expanding the coverage of relevant areas in the image.

\textbf{Representation injection in forward process.} When generating combinations of multiple subjects, directly adding image features and text features together, as depicted in \cref{eq:3}, can result in a loss of correspondence between the subjects and their attributes. Consequently, attribute mixing issues may arise, making it challenging to ensure consistency in the generated images with respect to the subjects.

To tackle this challenge and improve the alignment of subject attributes, we introduce a method of feature injection into the diffusion model after obtaining attention maps for each of the subjects. Specifically, for each subject $s_i$, we generate a mask based on its spatial feature map. Let $m_{i,j}, a_{i,j}$ in $M_{s}, A^{t}_{s} \in \mathbb{R}^{P \times P}$, respectively, represent the elements of the mask and attention map. The mask is calculated as follows:
\begin{equation}
m_{i,j} =
\begin{cases}
1, & \text{if } a_{i, j} > B \\
0, & \text{otherwise}
\end{cases}
\end{equation}
where $B$ denotes a threshold value, and we set it as $\frac{\max(A^{t}_{s})}{2}$. The resulting mask identifies the regions corresponding to the subject. Additionally, we introduce a forward guidance to update the subject image cross-attention output $z_{I}$ as follows:
\begin{equation}
z_{s_i}^{I} \leftarrow (1 - \mu)z_{s_i}^{I} + \mu M_{s_i} z_{s_{i}}^{I}
\label{eq:forward}
\end{equation}
In this equation, $\mu \in \left[ 0,1 \right]$ determines the strength of subject enhancement. The \cref{eq:3} is modified as
\begin{equation}
z \leftarrow z^{T} + \lambda \sum\limits_{s_{i}\in S}z_{s_i}^{I}
\label{eq:eq6}
\end{equation}
Notably, This guidance is injected during each forward denoising process and can be applied selectively across different resolution layers. For further details, please refer to the \cref{sec:appendixb}.

\textbf{Attention enhancement in backward process.} Given a text prompt consisting of $N$ tokens, we have $\sum_{i=1}^{N}A^{t}_{i} = 1$, and each token $y_{i}$ can be interpreted as competing for association with a specific location. Intuitively, subjects should have patches that distinctly attend to their corresponding tokens. Therefore, our optimization process aims to ensure that for each subject token $s_i$, there exists at least one patch in $A^{t}_{s}$ with a high attention value. As such, the loss function at timestep $t$ is defined as follows:
\begin{equation}
\mathcal{L} = 1 - \min\limits_{s_{i} \in S}  A^{t}_{s_i}
\label{eq:loss}
\end{equation}
To strengthen all subject tokens at the current timestep, the loss function maximizes the minimum value across the attention maps. In each denoising process, the latent variable $z_t$ is updated using backpropagation as follows:
\begin{equation}
z_{t} \leftarrow z_{t} - \eta \nabla_{z_t} \mathcal{L}
\label{eq:update}
\end{equation}
where $\eta$ is a scale factor controlling the strength of the guidance, changing dynamically over timestep $t$.

It is important to note that if the attention values of a token do not reach a certain threshold during the early denoising stages, the corresponding subject will not be generated. To address this, we have designed a loss threshold that is iteratively updated until the minimum attention value is achieved for all subject tokens. However, excessive updates can cause the latent variables to become out-of-distribution, leading to the production of incoherent images. To mitigate this, the refinement process is performed gradually across a subset of timesteps. 

Our proposed forward and backward guidance mechanisms are disentangled, allowing for independent or simultaneous utilization. Backward guidance improves the occupied region of the subject and facilitates better image injection. Furthermore, our guidance is applicable to both single and multiple object generation tasks. Multiple images of a single object can be inputted as a multi-perspective input, demonstrating an improvement in the fidelity of the generated subjects in the experiments.  \cref{sec:appendixb} provides additional details, including updates of the attention maps in the denoising process.

\section{Experiments}
\label{sec:experiments}

\subsection{Experimental Setup}
\textbf{Implementation details.} Our experiments use SD v1.5 \cite{stablediffusion} as the diffusion model and OpenCLIP ViT-H/14 \cite{Openclip} as the image encoder. The multi-modal prompts adapter is based on IP-Adapter \cite{ip-adapter}, which is trained on a multi-modal dataset consisting of 10 million text-image pairs from two publicly available datasets, LAION-2B \cite{laion} and COYO-700M, utilizing 8 V100 GPUs. Following these settings, we evaluate our method which does not require any additional training. In the inference stage, We generate 4 images for each text prompt and adopt a DDIM sampler with 50 steps, setting the guidance scale to 7.5 and $\mu$ to 0.8. Our method is generalizable to custom models fine-tuned from the SD v1.5 base model and we provide more details in the \cref{sec:appendixa}.

\begin{figure}[h]
  \centering
   \includegraphics[width=0.7\linewidth]{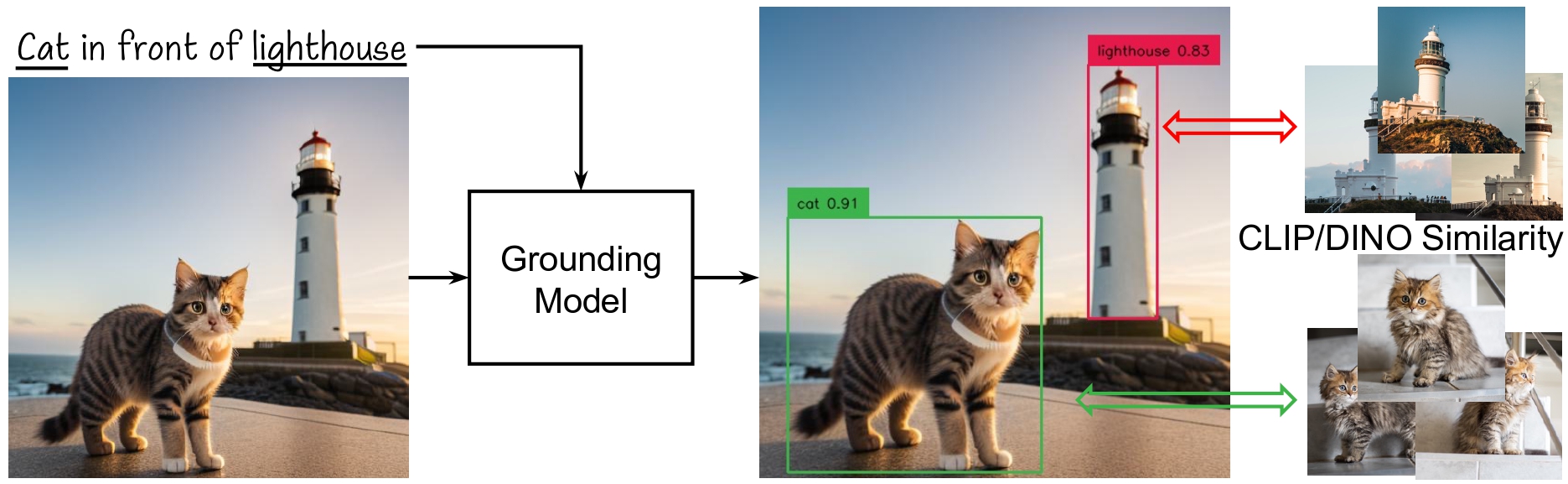}
   \caption{\textbf{GroundingScore metric.} The GroundingScore metric is calculated using the state-of-the-art Grounding model. In comparison to CLIP-I and DINO-I, GroundingScore provides a more accurate measurement of the fidelity of the subject image.} 
   \label{fig:groundingscore}
\end{figure}

\textbf{Datasets}
We perform experiments on CustomConcept101 \cite{customdiffusion}, consisting of 101 concepts with 3-15 images for each concept for evaluating model customization methods. 
ChatGPT \cite{ChatGPT} is used to generate 10-20 prompts for single and multiple concept generation.

\textbf{Evaluation metrics.} Prompt fidelity and subject consistency are essential factors for evaluating the quality of generated images. For the text alignment, we use CLIP-T: the text-image similarity between the text prompt and the generated images in CLIP feature space \cite{clipscore}. For the subject image alignment, we follow \cite{Dreambooth, TextualInversion, customdiffusion} and use CLIP-I and DINO-I: the similarity of generated images and source images embeddings extracted by CLIP \cite{CLIP} and DINO \cite{DINO}.

When evaluating subject-image alignment in compositional generation, it is crucial to focus on subject fidelity, which refers to the preservation of subject details in generated images. Therefore, we propose a new evaluation metric called GroundingScore to assess the task of compositional subject-driven generation. The objective of the Image Grounding task is to establish a visual connection between natural language descriptions and the corresponding regions or objects in an image. This objective aligns perfectly with our goal of achieving open-set object detection using text prompts.

\begin{figure*}[t]
  \centering
   \includegraphics[width=\linewidth]{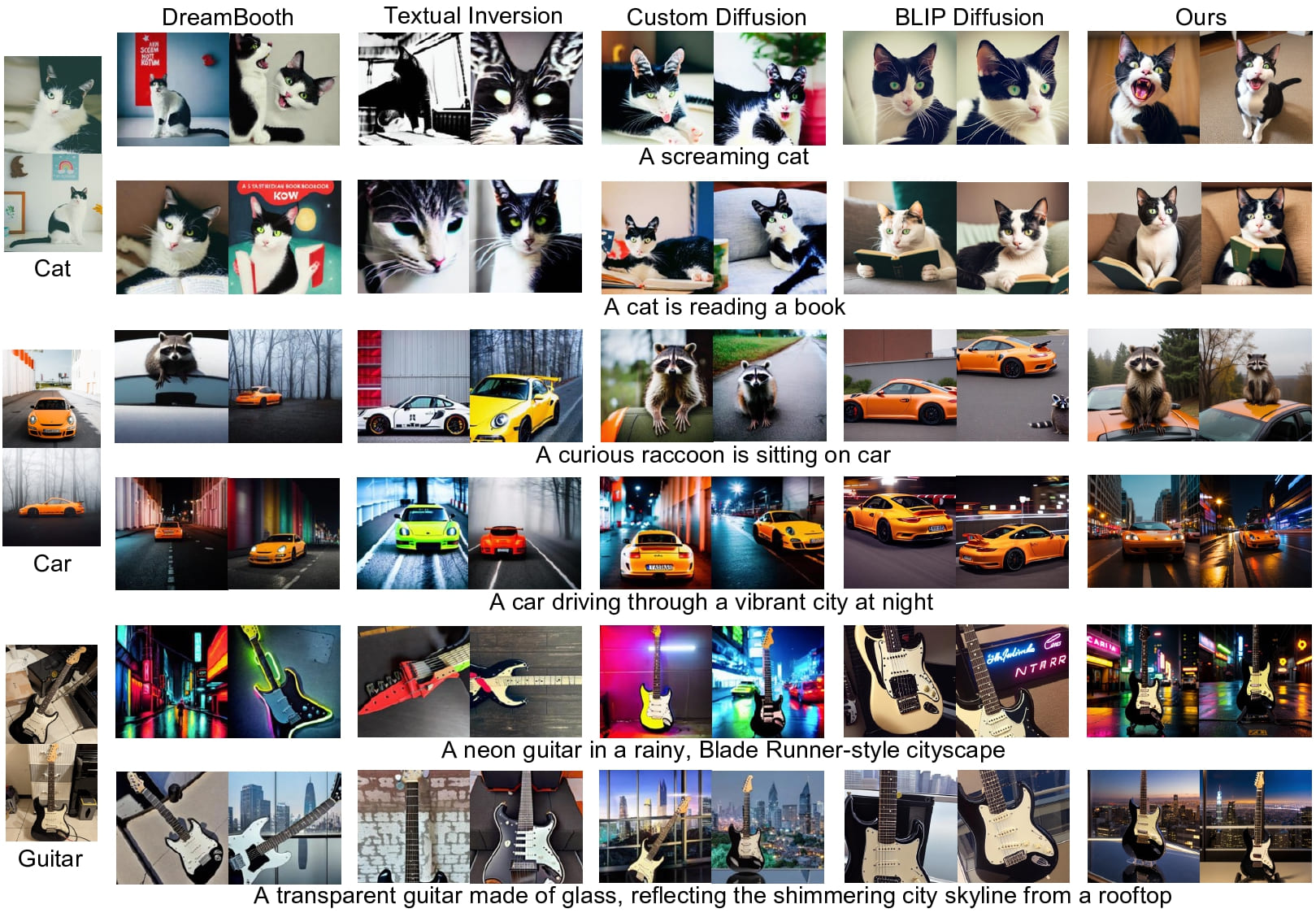}
   \caption{\textbf{Single-concept comparison results.} In cases where the text prompt is complex or entails the generation of multiple objects in conjunction, our zero-shot generations exhibit a subject-preservation performance that is comparable to that of the fine-tuned method, while demonstrating significantly enhanced image-text alignment when compared to other methods.} 
   \label{fig:compare-single}
\end{figure*}

\cref{fig:groundingscore} shows the calculation way of the GroundingScore. Specifically, we employ a state-of-the-art grounding model \cite{groundingdino, GLIP, GLIPv2, detclip, VILD, Regionclip} to detect the regions where the target subjects are located. We then extract features from the grounded regions and calculate the CLIP-I and DINO-I scores. If a particular subject is not detected, its score is assigned as 0. Finally, we computed the average results for multiple subjects. The final results were defined as CLIP-GroundingScore (CLIP-GS) when employing CLIP as the image encoder, and DINO-GroundingScore (DINO-GS) when employing DINO as the image encoder. In our experiments, we utilize GroundingDINO-T \cite{groundingdino} as an image backbone and BERT-base \cite{BERT} as the text backbone. We set the box threshold to 0.35 and the text threshold to 0.25.

\subsection{Comparison Experiments}

\begin{figure}[t]
\centering
\begin{minipage}[b]{0.49\linewidth}
\includegraphics[width=\linewidth]{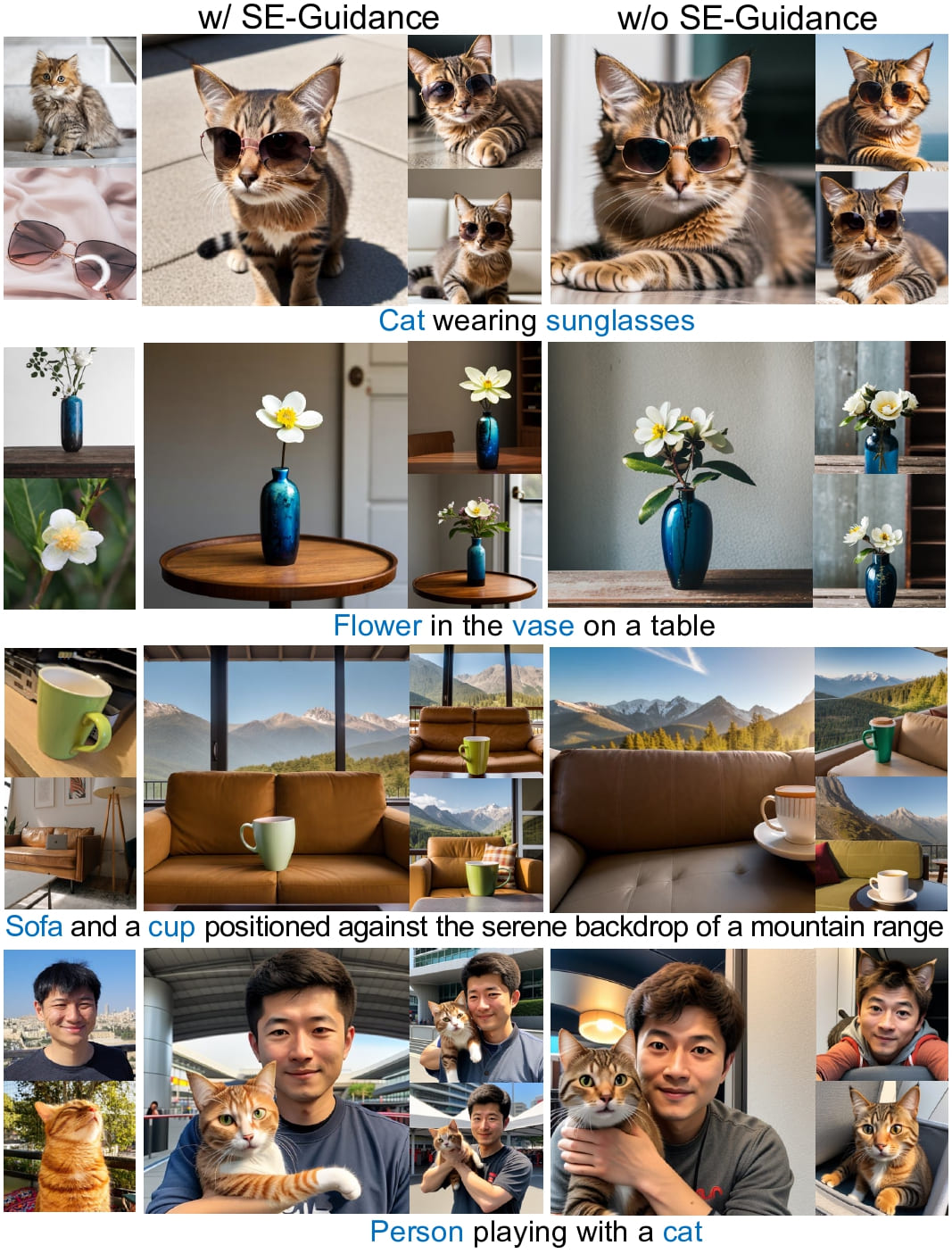}
\caption{\textbf{Multi-concepts comparison results.} Our SE-Guidance exhibits exceptional image-text alignment and enhances the consistency of subject images, particularly in understanding the combination relationship between two subjects and generating smaller objects.}
\label{fig:multicases}
\end{minipage}
\hfill
\begin{minipage}[b]{0.49\linewidth}
\includegraphics[width=\linewidth]{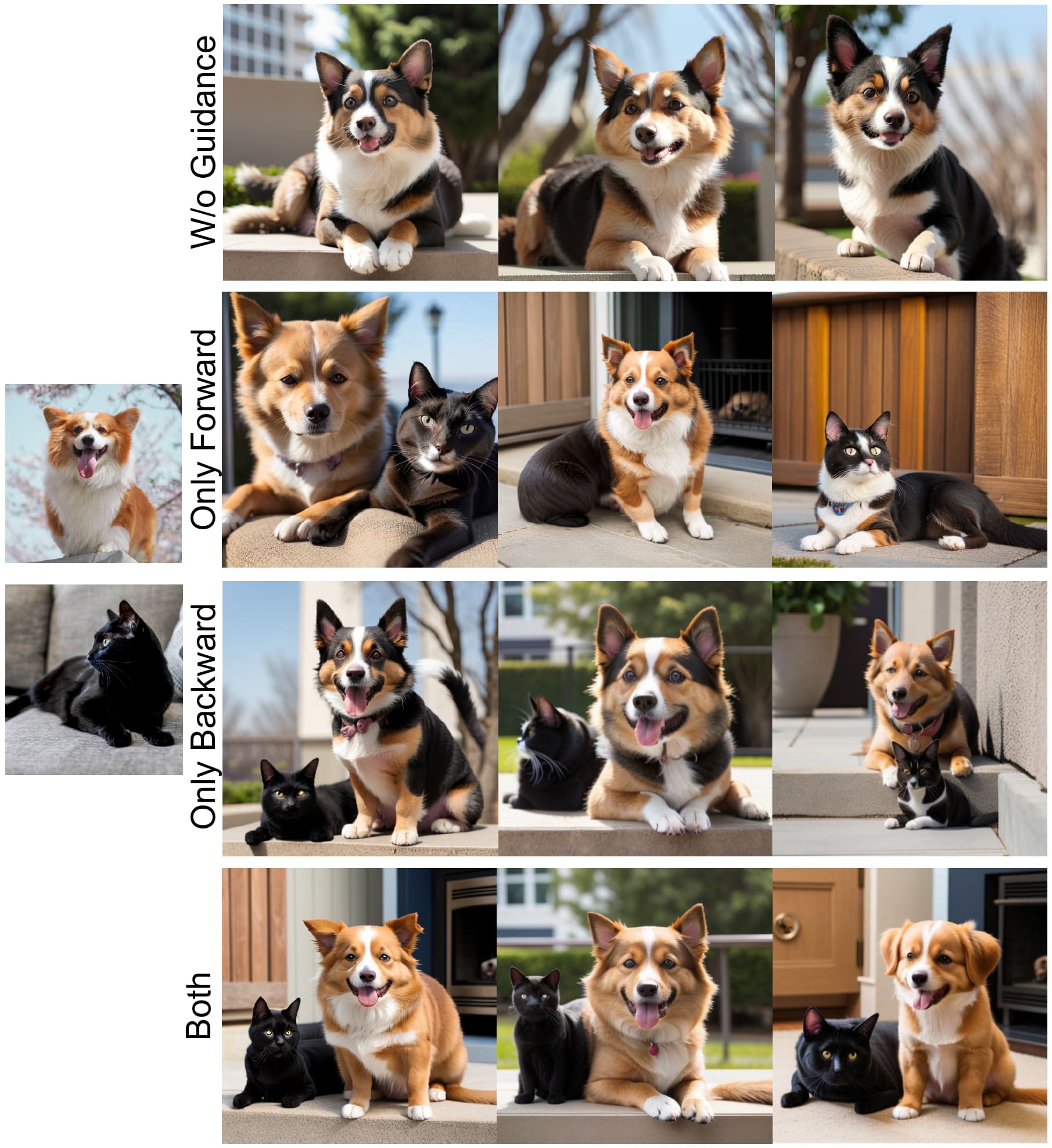}
\caption{\textbf{Forward guidance vs. Backward guidance.} The textual prompt is "A dog and a cat". Forward guidance helps distinguish subject attributes, but accurate feature injection is not achieved. Backward guidance effectively mitigates subject loss but struggles to ensure fidelity for each subject.}
\label{fig:ablation}
\end{minipage}
\end{figure}

\textbf{Single concept generation.} We compare our method with three fine-tune-based methods: DreamBooth \cite{Dreambooth}, Textual Inversion \cite{TextualInversion}, Custom Diffusion \cite{customdiffusion}, and a zero-shot method BLIP-Diffusion \cite{Blipdiffusion}. DreamBooth fine-tunes all the parameters in the diffusion model while keeping the text transformer frozen, and Textual Inversion optimizes a new placeholder token for each new concept. Custom Diffusion updates $W_k$ and $W_v$ in the cross-attention block and other parameters are frozen. BLIP-Diffusion is a zero-shot subject-driven generation model using subject representation learning.

\cref{tab:compare} shows the quantitative comparison results of our methods with other methods. Within the fine-tune-based methods, we observe that finetuning more parameters enhances the ability to maintain subject identity, with DreamBooth achieving the highest score in this regard. Zero-shot BLIP-Diffusion achieves the highest image-image alignment. However, we observe a significant decrease in its text-image alignment, which can be attributed to overfitting the input images. Our zero-shot generations display a comparable subject-preservation performance compared with the fine-tune-based method while showing significant image-text alignment, particularly in cases where the text prompt is complex or requires the generation of multiple objects together, as depicted in \cref{fig:compare-single}. Additionally, our method supports multi-view image inputs for single-subject generation, we observe improvements in all performance metrics when an additional image is added as input.

\textbf{Compositional concepts generation.} We compare our method with Custom Diffusion and directly modify the IP-Adapter by summing cross-attention block outputs of each subject as our baselines. Our method achieves comparable results to Custom Diffusion, which requires multiple images per subject and fine-tuning. This demonstrates the feasibility and superiority of our zero-shot approach, as our method only requires providing one image per subject. Additional analysis and examples are presented in \cref{sec:appendixc}.

\begin{table*}[t]
    \centering
  \caption{\textbf{Quantitative comparisons.} We compare our method with fine-tune-based and zero-shot methods on single-concept and multi-concept subject-driven generation on CustomConcept101 \cite{customdiffusion} dataset. FT means fine-tuning (Y) or not (N). CLIP-GroundingScore (CLIP-GS) employs CLIP as the image encoder and DINO-GroundingScore (DINO-GS) employs DINO as the image encoder.}
  \label{tab:compare}
    \fontsize{8pt}{10pt}\selectfont
    \begin{tabular}{p{0.8cm} l c c c c c c c}
    \toprule
         & \textbf{Method} & \textbf{FT} & \textbf{CLIP-T} & \textbf{CLIP-I} & \textbf{DINO-I} & \textbf{CLIP-GS} & \textbf{DINO-GS} \\ 
         \midrule
        \multirow{6}{*}{Single} & DreamBooth \cite{Dreambooth} & Y & 0.7533 & 0.7533 & 0.5557    & 0.6528 & 0.4724 & \\ 
        ~ & Textual Inversion \cite{TextualInversion} & Y & 0.6128 & 0.7483 & 0.5115 & 0.6124 & 0.4073 \\ 
        ~ & Custom Diffusion \cite{customdiffusion} & Y & \textbf{0.7719} & 0.7475 & 0.5371 & 0.6394 & 0.4511 \\ 
        ~ & BLIP-Diffusion \cite{Blipdiffusion}& N & 0.6673 & \textbf{0.7847} & \textbf{0.5845} & \textbf{0.6657} & \textbf{0.4814} \\ 
        ~ & IP-Adapter \cite{ip-adapter} & N & 0.7580 & 0.7446 & 0.5244 & 0.6371 & 0.4344 \\ 
        ~ &  + SE-Guidance & N & 0.7435 & 0.7552 & 0.5344 & 0.6422 & 0.4432 \\    
        ~ &  + Multi-view & N & 0.7455 & 0.7582 & 0.5400 & 0.6447 & 0.4452 \\   
        \midrule
        \multirow{3}{*}{Multi} & Custom Diffusion \cite{customdiffusion}& Y & 0.7506 & 0.6687 & \textbf{0.3779} & 0.6136 & 0.3508 \\
        ~ & IP-Adapter\textsuperscript{*} \cite{ip-adapter} & N & 0.7250 & \textbf{0.6858}    & 0.3711 &    0.6012    & 0.3454 \\ 
        ~ &  + SE-Guidance & N & \textbf{0.7714} &0.6751 &0.3648&    \textbf{0.6391}&   \textbf{0.3632} \\ 
        \bottomrule
    \end{tabular}
\end{table*}

As shown in \cref{fig:multicases}, our method demonstrates a significant superiority in image-text alignment. Additionally, we also observe an improvement in the consistency of generated subject images after incorporating SE-Guidance. However, CLIP-I and DINO-I do not effectively demonstrate this enhancement. We believe this could be attributed to the relatively smaller regions occupied by each subject in the generated composite images, thereby affecting their similarity to the original image. Our proposed GroundingScore validates this observation by showcasing improved metrics, as depicted in \cref{tab:compare}.

\begin{table}[htb]
  \caption{\textbf{User Study.} "Alignment" and "Fidelity" represent the vision-language alignment and the identity preservation performance.}
  \label{tab:userstudy}
\centering
\begin{tabular}{c c c c c}
\toprule
\multirow{2}{*}{Method} & \multicolumn{2}{c}{Single Subject} & \multicolumn{2}{c}{Multi Subjects} \\ \cmidrule(lr){2-3} \cmidrule(lr){4-5}
               & Alignment    & Fidelity    & Alignment    & Fidelity   \\ 
\midrule
Dreambooth\cite{Dreambooth} &     3.03     &      3.31     &     -      &   -  \\
Textual Inversion\cite{TextualInversion} &     2.75     &    2.69       &    -      &  -   \\
BLIP Diffusion\cite{Blipdiffusion} &    2.31    &    \textbf{3.57}        &      -    &    -    \\
CustomDiffusion\cite{customdiffusion} &     3.18     &    3.27        &     2.67      &  3.03      \\
IP-Adapter\cite{ip-adapter}      &    \textbf{3.95}     &    3.18         &      3.35    &   2.87     \\
Ours            &     3.91      &     3.38        &      \textbf{3.52}      &  \textbf{3.14}      \\
\bottomrule
\end{tabular}
\end{table}

\textbf{User Study.} We conduct a user study to compare our method with other methods. We randomly select 20 subjects and generate 200 images for each model using 5 corresponding text prompts and 2 random seeds. Three users rank each generated image from 1 (worst) to 5 (best) based on the generated image's language-vision alignment and preservation of identity. The mean score for individual models was computed and is displayed in \cref{tab:userstudy}. Images generated through BLIP-Diffusion are prone to overfitting, resulting in high subject consistency but failure to adhere to the semantics of the text prompts. Our method performs the best except for image alignment in the single subject generation. This correlates with our quantitative assessment.

\subsection{Ablation Study}

\textbf{Forward and backward guidance.} The effectiveness of each module was verified by individually applying different operations in the compositional generation task, as our proposed forward and backward guidance mechanisms are disentangled. As shown in \cref{fig:ablation}, the results without guidance exhibited a loss of subject clarity, with noticeable attribute mixing between the two objects. Solely employing forward guidance helped distinguish the attributes of the two subjects but failed to achieve accurate feature injection. On the other hand, solely applying backward guidance effectively mitigated the issue of subject loss, yet struggled to ensure fidelity for each subject. Finally, the simultaneous integration of forward and backward guidance yielded the best results, significantly restoring the features of both subjects. 

\begin{figure}[htb]
    \begin{minipage}[b]{0.52\linewidth}
        \centering
        \includegraphics[width=\linewidth]{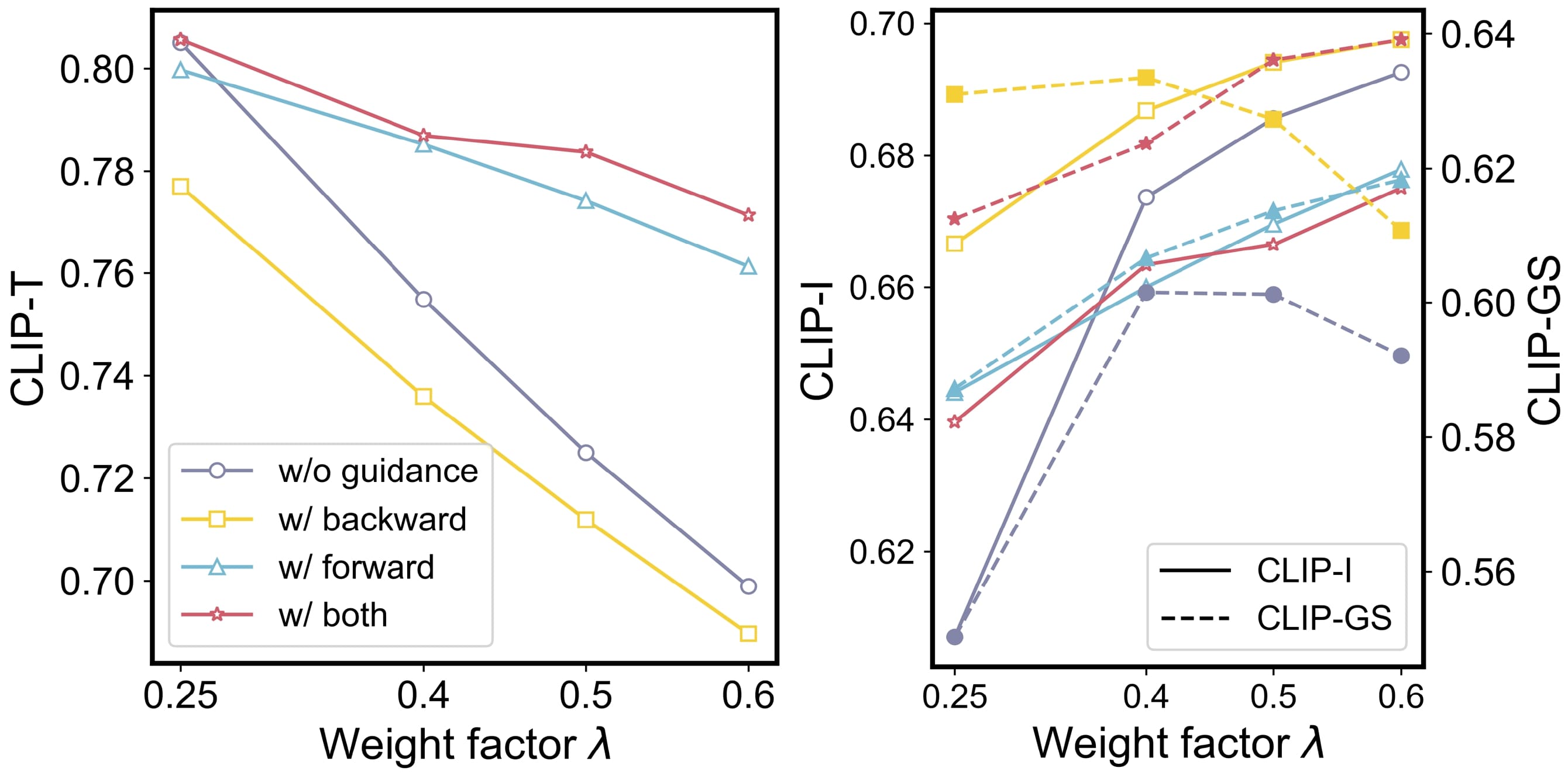}
        \caption{\textbf{Ablation study of weight factor.} Our method maintains a high level of subject fidelity with only a minimal decline in text-image consistency as the weight factor $\lambda$ increases.}
        \label{fig:line_figure}
    \end{minipage}
    \hfill
    \begin{minipage}[b]{0.47\linewidth}
    \centering
   \includegraphics[width=\linewidth]{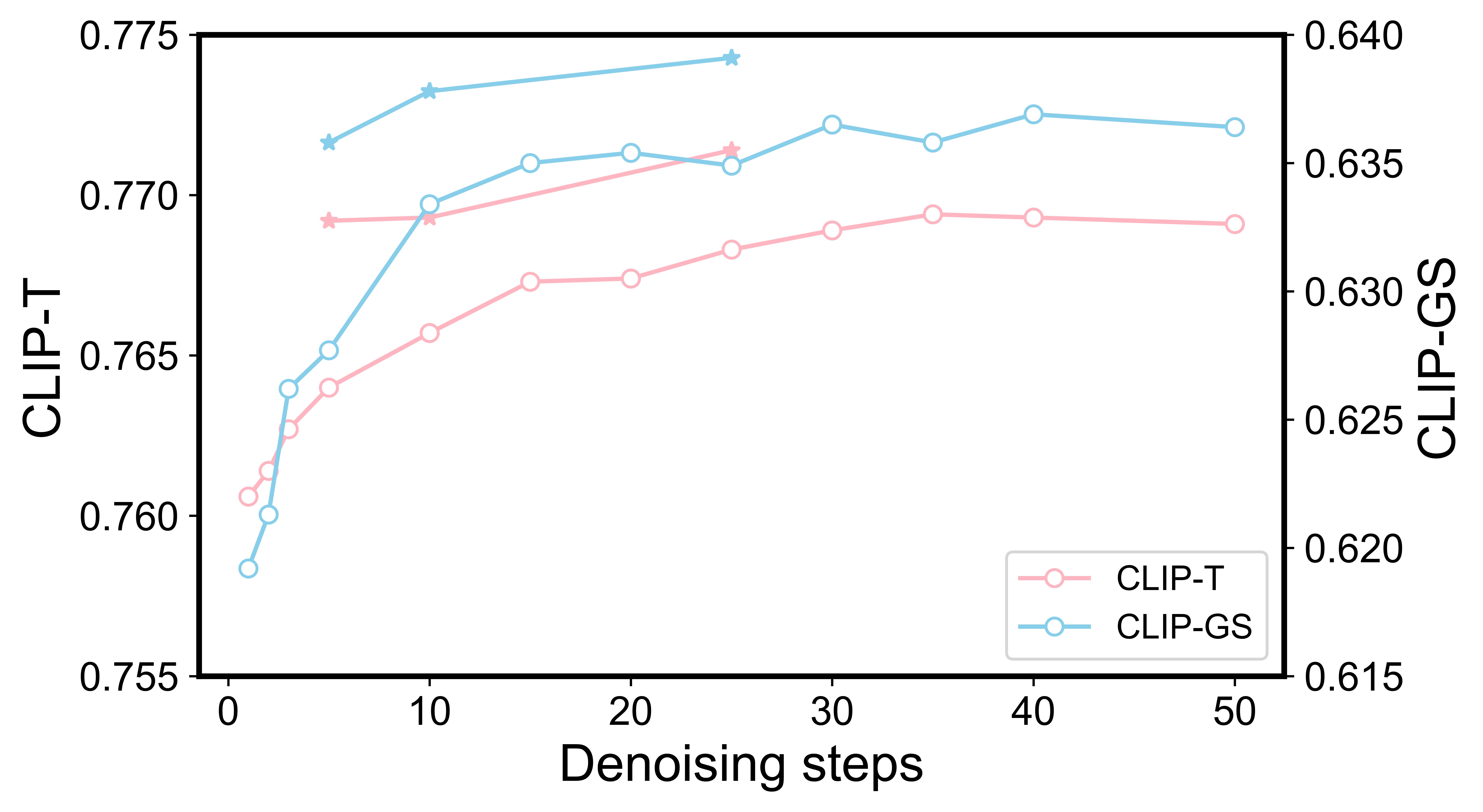}
   \caption{\textbf{Ablation study on the denoising steps of SE-Guidance.} $\circ$ and $\star$ represent the experimental results for without and with refinement processes.}
   \label{fig:steps}
    \end{minipage}
\end{figure}

Furthermore, in addition to the qualitative comparison of the forward and backward mechanisms, we conducted a quantitative analysis as well, which is presented in \cref{tab:ablation} and \cref{fig:line_figure}. This quantitative analysis consistently reveals that the simultaneous application of both modules leads to the highest scores for both CLIP-T and GroundingScore, illustrating exceptional semantic coherence and subject fidelity.

\textbf{Strength of image prompt.} We conduct ablation studies to analyze the weight factor $\lambda$, which controls the strength of image feature injection. As shown in \cref{fig:line_figure}, the image-text alignment of the IP-Adapter rapidly deteriorates as $\lambda$ increases, resulting in generated images that deviate from the influence of the text prompt and closely resemble the original image. Interestingly, this leads to an improvement in DINO-I and CLIP-I scores. Moreover, the GroundingScore metric indicates that the inclusion of Se-Guidance enhances image alignment. Notably, our method is not highly sensitive to changes in $\lambda$, as it maintains a high subject fidelity while only experiencing a minimal decline in text-image consistency. This enables us to generate more diverse subject images by adjusting the intensity of the image input.

\begin{table}[htb]
\captionof{table}{\textbf{Ablation results.} Fw and Bw mean whether to perform forward and backward guidance. It is evident from the metrics of CLIP-T and CLIP-GS that our method demonstrates exceptional semantic coherence and subject fidelity.}
\label{tab:ablation}
\centering
\begin{tabular}{c c c c c c}
    \toprule
    Fw & Bw & CLIP-T & CLIP-I & CLIP-GS & Speed \\ 
    \midrule
    ~ & ~ & 0.6989 & 0.6926 & 0.5921 &  2 s/img \\      
    \checkmark & & 0.7742 & 0.6696 & 0.6137 & 6 s/img \\ 
    & \checkmark & 0.7119 & \textbf{0.6941} & 0.6273 & 8 s/img\\ 
    \checkmark  & \checkmark & \textbf{0.7837} & 0.6665 & \textbf{0.6361} & 13 s/img\\ 
    \bottomrule
\end{tabular}
\end{table}

\textbf{Inference speed.} After implementing the SE-Guidance, we observed the speed of each DDIM denoising step for generating one image decreased from 0.04s to 0.24s on one NVIDIA 32G V100. Then, we attempted to employ guidance exclusively during the initial steps, aiming to strike a balance between time consumption and object-attribute binding. We perform guidance for the first several timesteps respectively in a 50-step DDIM, and the results demonstrate that incorporating guidance solely in the preceding steps plays a more crucial role in facilitating attribute binding. Moreover, we have validated the effectiveness of the refinement process, particularly in significantly enhancing the generation of the subject during the initial denoising steps as shown in \cref{fig:steps}.  Significant improvements in performance can be achieved by setting the attention threshold at the 5th step (0.5) and 10th step (0.8).

The forward process introduces a computational burden as it involves recording and multiplying attention maps for each timestep, and the backward guidance requires gradient backpropagation. We calculate the speed of the SE-guidance with a 10-step refinement process in \cref{fig:steps}. Notably, in terms of the total time required for training and inference, our method is significantly faster than the fine-tune-based methods like CustomDiffusion($\sim$10min), Dreambooth($\sim$25min), Textual Inversion($\sim$1h) for each instance on two V100s. 
\section{Limitations}
\label{sec:limitations}

\begin{figure}[htb]
  \centering
   \includegraphics[width=0.70\linewidth]{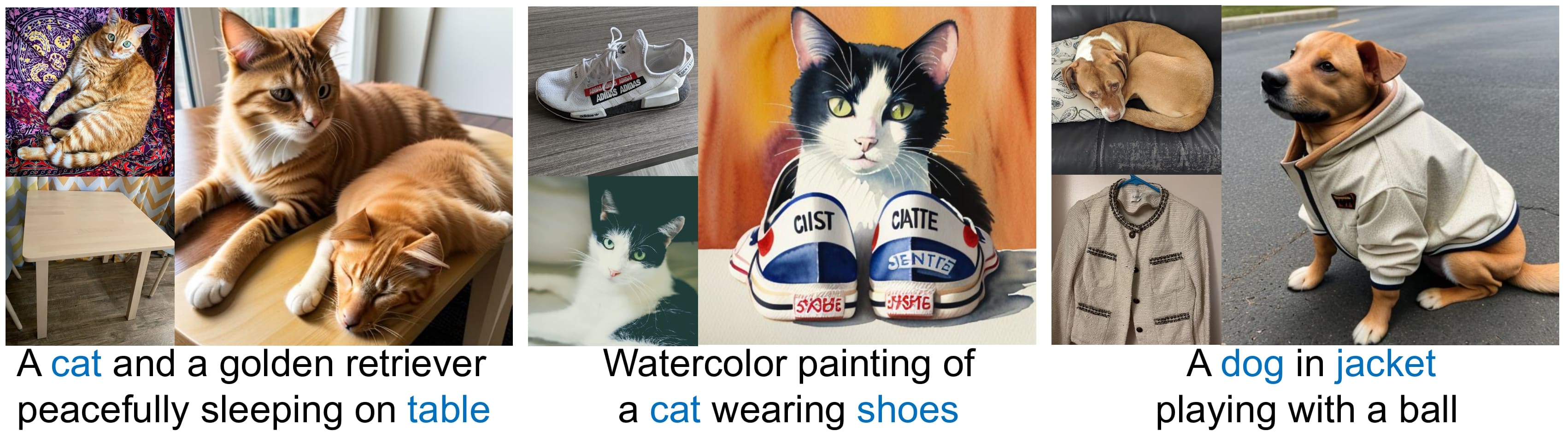}
   \caption{\textbf{Failure cases on compositional generation.} Our method encounters challenges in generating fine-grained composition relations, and the effectiveness is constrained by the expressive capacity of the generative model.} 
   \label{fig:failure}
\end{figure}

Our zero-shot method is constrained by the expressive power of the generative model as we do not utilize additional training. Some failure cases are shown in \cref{fig:failure}. When dealing with unique and rare objects, it becomes challenging to ensure high consistency in details. Furthermore, when attempting to synthesize subjects that naturally do not appear together, the generated images may lack realism. Additionally, addressing fine-grained composition relations is also a daunting task, particularly when more than two subjects are involved, as it can lead to issues like attribute mixing or missing.

\section{Conclusions}
\label{sec:conclusions}

In conclusion, we propose a straightforward and effective method for improving subject-driven text-to-image generation, especially compositional generation tasks. We apply training-free guidance that leverages cross-attention maps to bind subject image features with their corresponding semantic information while disentangling subject pairs from each other. Additionally, we introduce new evaluation metrics for a more comprehensive assessment of the task. In future work, we plan to explore a more versatile approach to incorporating image features in generation tasks, allowing for better alignment between images and textual prompts.

%
%
\bibliographystyle{splncs04}
\bibliography{main}

\begin{thebibliography}{10}
\providecommand{\url}[1]{\texttt{#1}}
\providecommand{\urlprefix}{URL }
\providecommand{\doi}[1]{https://doi.org/#1}

\bibitem{cosposition3}
Ashual, O., Wolf, L.: Specifying object attributes and relations in interactive scene generation. In: Proceedings of the IEEE/CVF international conference on computer vision. pp. 4561--4569 (2019)

\bibitem{ediffi}
Balaji, Y., Nah, S., Huang, X., Vahdat, A., Song, J., Kreis, K., Aittala, M., Aila, T., Laine, S., Catanzaro, B., et~al.: ediffi: Text-to-image diffusion models with an ensemble of expert denoisers. arXiv preprint arXiv:2211.01324  (2022)

\bibitem{Uvit}
Bao, F., Nie, S., Xue, K., Cao, Y., Li, C., Su, H., Zhu, J.: All are worth words: A vit backbone for diffusion models. In: Proceedings of the IEEE/CVF Conference on Computer Vision and Pattern Recognition. pp. 22669--22679 (2023)

\bibitem{biggan}
Brock, A., Donahue, J., Simonyan, K.: Large scale gan training for high fidelity natural image synthesis. arXiv preprint arXiv:1809.11096  (2018)

\bibitem{DINO}
Caron, M., Touvron, H., Misra, I., J{\'e}gou, H., Mairal, J., Bojanowski, P., Joulin, A.: Emerging properties in self-supervised vision transformers. In: Proceedings of the IEEE/CVF international conference on computer vision. pp. 9650--9660 (2021)

\bibitem{attendandexcite}
Chefer, H., Alaluf, Y., Vinker, Y., Wolf, L., Cohen-Or, D.: Attend-and-excite: Attention-based semantic guidance for text-to-image diffusion models. ACM Transactions on Graphics (TOG)  \textbf{42}(4),  1--10 (2023)

\bibitem{disenbooth}
Chen, H., Zhang, Y., Wang, X., Duan, X., Zhou, Y., Zhu, W.: Disenbooth: Disentangled parameter-efficient tuning for subject-driven text-to-image generation. arXiv preprint arXiv:2305.03374  (2023)

\bibitem{layoutguidance}
Chen, M., Laina, I., Vedaldi, A.: Training-free layout control with cross-attention guidance. arXiv preprint arXiv:2304.03373  (2023)

\bibitem{SuTI}
Chen, W., Hu, H., Li, Y., Rui, N., Jia, X., Chang, M.W., Cohen, W.W.: Subject-driven text-to-image generation via apprenticeship learning. arXiv preprint arXiv:2304.00186  (2023)

\bibitem{re-imagen}
Chen, W., Hu, H., Saharia, C., Cohen, W.W.: Re-imagen: Retrieval-augmented text-to-image generator. arXiv preprint arXiv:2209.14491  (2022)

\bibitem{Openclip}
Cherti, M., Beaumont, R., Wightman, R., Wortsman, M., Ilharco, G., Gordon, C., Schuhmann, C., Schmidt, L., Jitsev, J.: Reproducible scaling laws for contrastive language-image learning. In: Proceedings of the IEEE/CVF Conference on Computer Vision and Pattern Recognition. pp. 2818--2829 (2023)

\bibitem{zestguide}
Couairon, G., Careil, M., Cord, M., Lathuili{\`e}re, S., Verbeek, J.: Zero-shot spatial layout conditioning for text-to-image diffusion models. In: Proceedings of the IEEE/CVF International Conference on Computer Vision. pp. 2174--2183 (2023)

\bibitem{BERT}
Devlin, J., Chang, M.W., Lee, K., Toutanova, K.: Bert: Pre-training of deep bidirectional transformers for language understanding. arXiv preprint arXiv:1810.04805  (2018)

\bibitem{tamingtransformer}
Esser, P., Rombach, R., Ommer, B.: Taming transformers for high-resolution image synthesis. In: Proceedings of the IEEE/CVF conference on computer vision and pattern recognition. pp. 12873--12883 (2021)

\bibitem{structurediffusion}
Feng, W., He, X., Fu, T.J., Jampani, V., Akula, A., Narayana, P., Basu, S., Wang, X.E., Wang, W.Y.: Training-free structured diffusion guidance for compositional text-to-image synthesis. arXiv preprint arXiv:2212.05032  (2022)

\bibitem{MakeAScene}
Gafni, O., Polyak, A., Ashual, O., Sheynin, S., Parikh, D., Taigman, Y.: Make-a-scene: Scene-based text-to-image generation with human priors. In: European Conference on Computer Vision. pp. 89--106. Springer (2022)

\bibitem{TextualInversion}
Gal, R., Alaluf, Y., Atzmon, Y., Patashnik, O., Bermano, A.H., Chechik, G., Cohen-Or, D.: An image is worth one word: Personalizing text-to-image generation using textual inversion. arXiv preprint arXiv:2208.01618  (2022)

\bibitem{designing}
Gal, R., Arar, M., Atzmon, Y., Bermano, A.H., Chechik, G., Cohen-Or, D.: Designing an encoder for fast personalization of text-to-image models. arXiv preprint arXiv:2302.12228  (2023)

\bibitem{vqdiff}
Gu, S., Chen, D., Bao, J., Wen, F., Zhang, B., Chen, D., Yuan, L., Guo, B.: Vector quantized diffusion model for text-to-image synthesis. In: Proceedings of the IEEE/CVF Conference on Computer Vision and Pattern Recognition. pp. 10696--10706 (2022)

\bibitem{VILD}
Gu, X., Lin, T.Y., Kuo, W., Cui, Y.: Open-vocabulary object detection via vision and language knowledge distillation. arXiv preprint arXiv:2104.13921  (2021)

\bibitem{svdiff}
Han, L., Li, Y., Zhang, H., Milanfar, P., Metaxas, D., Yang, F.: Svdiff: Compact parameter space for diffusion fine-tuning. arXiv preprint arXiv:2303.11305  (2023)

\bibitem{prompt2prompt}
Hertz, A., Mokady, R., Tenenbaum, J., Aberman, K., Pritch, Y., Cohen-Or, D.: Prompt-to-prompt image editing with cross attention control. arXiv preprint arXiv:2208.01626  (2022)

\bibitem{clipscore}
Hessel, J., Holtzman, A., Forbes, M., Bras, R.L., Choi, Y.: Clipscore: A reference-free evaluation metric for image captioning. arXiv preprint arXiv:2104.08718  (2021)

\bibitem{DDPM}
Ho, J., Jain, A., Abbeel, P.: Denoising diffusion probabilistic models. Advances in neural information processing systems  \textbf{33},  6840--6851 (2020)

\bibitem{amtgan}
Hu, S., Liu, X., Zhang, Y., Li, M., Zhang, L.Y., Jin, H., Wu, L.: Protecting facial privacy: Generating adversarial identity masks via style-robust makeup transfer. In: Proceedings of the IEEE/CVF Conference on Computer Vision and Pattern Recognition. pp. 15014--15023 (2022)

\bibitem{GAN}
Isola, P., Zhu, J.Y., Zhou, T., Efros, A.A.: Image-to-image translation with conditional adversarial networks. In: Proceedings of the IEEE conference on computer vision and pattern recognition. pp. 1125--1134 (2017)

\bibitem{tamingencoder}
Jia, X., Zhao, Y., Chan, K.C., Li, Y., Zhang, H., Gong, B., Hou, T., Wang, H., Su, Y.C.: Taming encoder for zero fine-tuning image customization with text-to-image diffusion models. arXiv preprint arXiv:2304.02642  (2023)

\bibitem{pggan}
Karras, T., Aila, T., Laine, S., Lehtinen, J.: Progressive growing of gans for improved quality, stability, and variation. arXiv preprint arXiv:1710.10196  (2017)

\bibitem{Stylegan}
Karras, T., Laine, S., Aila, T.: A style-based generator architecture for generative adversarial networks. In: Proceedings of the IEEE/CVF conference on computer vision and pattern recognition. pp. 4401--4410 (2019)

\bibitem{gan4}
Karras, T., Laine, S., Aittala, M., Hellsten, J., Lehtinen, J., Aila, T.: Analyzing and improving the image quality of stylegan. In: Proceedings of the IEEE/CVF conference on computer vision and pattern recognition. pp. 8110--8119 (2020)

\bibitem{customdiffusion}
Kumari, N., Zhang, B., Zhang, R., Shechtman, E., Zhu, J.Y.: Multi-concept customization of text-to-image diffusion. In: Proceedings of the IEEE/CVF Conference on Computer Vision and Pattern Recognition. pp. 1931--1941 (2023)

\bibitem{Blipdiffusion}
Li, D., Li, J., Hoi, S.C.: Blip-diffusion: Pre-trained subject representation for controllable text-to-image generation and editing. arXiv preprint arXiv:2305.14720  (2023)

\bibitem{GLIP}
Li, L.H., Zhang, P., Zhang, H., Yang, J., Li, C., Zhong, Y., Wang, L., Yuan, L., Zhang, L., Hwang, J.N., et~al.: Grounded language-image pre-training. In: Proceedings of the IEEE/CVF Conference on Computer Vision and Pattern Recognition. pp. 10965--10975 (2022)

\bibitem{PNDM}
Liu, L., Ren, Y., Lin, Z., Zhao, Z.: Pseudo numerical methods for diffusion models on manifolds. arXiv preprint arXiv:2202.09778  (2022)

\bibitem{composablediffusion}
Liu, N., Li, S., Du, Y., Torralba, A., Tenenbaum, J.B.: Compositional visual generation with composable diffusion models. In: European Conference on Computer Vision. pp. 423--439. Springer (2022)

\bibitem{groundingdino}
Liu, S., Zeng, Z., Ren, T., Li, F., Zhang, H., Yang, J., Li, C., Yang, J., Su, H., Zhu, J., et~al.: Grounding dino: Marrying dino with grounded pre-training for open-set object detection. arXiv preprint arXiv:2303.05499  (2023)

\bibitem{Cones}
Liu, Z., Feng, R., Zhu, K., Zhang, Y., Zheng, K., Liu, Y., Zhao, D., Zhou, J., Cao, Y.: Cones: Concept neurons in diffusion models for customized generation. arXiv preprint arXiv:2303.05125  (2023)

\bibitem{Cones2}
Liu, Z., Zhang, Y., Shen, Y., Zheng, K., Zhu, K., Feng, R., Liu, Y., Zhao, D., Zhou, J., Cao, Y.: Cones 2: Customizable image synthesis with multiple subjects. arXiv preprint arXiv:2305.19327  (2023)

\bibitem{DPMSolver}
Lu, C., Zhou, Y., Bao, F., Chen, J., Li, C., Zhu, J.: Dpm-solver: A fast ode solver for diffusion probabilistic model sampling in around 10 steps. Advances in Neural Information Processing Systems  \textbf{35},  5775--5787 (2022)

\bibitem{t2i-adapter}
Mou, C., Wang, X., Xie, L., Zhang, J., Qi, Z., Shan, Y., Qie, X.: T2i-adapter: Learning adapters to dig out more controllable ability for text-to-image diffusion models. arXiv preprint arXiv:2302.08453  (2023)

\bibitem{glide}
Nichol, A., Dhariwal, P., Ramesh, A., Shyam, P., Mishkin, P., McGrew, B., Sutskever, I., Chen, M.: Glide: Towards photorealistic image generation and editing with text-guided diffusion models. arXiv preprint arXiv:2112.10741  (2021)

\bibitem{ChatGPT}
Ouyang, L., Wu, J., Jiang, X., Almeida, D., Wainwright, C., Mishkin, P., Zhang, C., Agarwal, S., Slama, K., Ray, A., et~al.: Training language models to follow instructions with human feedback. Advances in Neural Information Processing Systems  \textbf{35},  27730--27744 (2022)

\bibitem{sdxl}
Podell, D., English, Z., Lacey, K., Blattmann, A., Dockhorn, T., M{\"u}ller, J., Penna, J., Rombach, R.: Sdxl: improving latent diffusion models for high-resolution image synthesis. arXiv preprint arXiv:2307.01952  (2023)

\bibitem{qiao2019learn}
Qiao, T., Zhang, J., Xu, D., Tao, D.: Learn, imagine and create: Text-to-image generation from prior knowledge. Advances in neural information processing systems  \textbf{32} (2019)

\bibitem{CLIP}
Radford, A., Kim, J.W., Hallacy, C., Ramesh, A., Goh, G., Agarwal, S., Sastry, G., Askell, A., Mishkin, P., Clark, J., et~al.: Learning transferable visual models from natural language supervision. In: International conference on machine learning. pp. 8748--8763. PMLR (2021)

\bibitem{dcgan}
Radford, A., Metz, L., Chintala, S.: Unsupervised representation learning with deep convolutional generative adversarial networks. arXiv preprint arXiv:1511.06434  (2015)

\bibitem{DALLE2}
Ramesh, A., Dhariwal, P., Nichol, A., Chu, C., Chen, M.: Hierarchical text-conditional image generation with clip latents. arXiv preprint arXiv:2204.06125  \textbf{1}(2), ~3 (2022)

\bibitem{vqvae2}
Razavi, A., Van~den Oord, A., Vinyals, O.: Generating diverse high-fidelity images with vq-vae-2. Advances in neural information processing systems  \textbf{32} (2019)

\bibitem{kandinsky}
Razzhigaev, A., Shakhmatov, A., Maltseva, A., Arkhipkin, V., Pavlov, I., Ryabov, I., Kuts, A., Panchenko, A., Kuznetsov, A., Dimitrov, D.: Kandinsky: an improved text-to-image synthesis with image prior and latent diffusion. arXiv preprint arXiv:2310.03502  (2023)

\bibitem{stablediffusion}
Rombach, R., Blattmann, A., Lorenz, D., Esser, P., Ommer, B.: High-resolution image synthesis with latent diffusion models. In: Proceedings of the IEEE/CVF conference on computer vision and pattern recognition. pp. 10684--10695 (2022)

\bibitem{Dreambooth}
Ruiz, N., Li, Y., Jampani, V., Pritch, Y., Rubinstein, M., Aberman, K.: Dreambooth: Fine tuning text-to-image diffusion models for subject-driven generation. In: Proceedings of the IEEE/CVF Conference on Computer Vision and Pattern Recognition. pp. 22500--22510 (2023)

\bibitem{imagen}
Saharia, C., Chan, W., Saxena, S., Li, L., Whang, J., Denton, E.L., Ghasemipour, K., Gontijo~Lopes, R., Karagol~Ayan, B., Salimans, T., et~al.: Photorealistic text-to-image diffusion models with deep language understanding. Advances in Neural Information Processing Systems  \textbf{35},  36479--36494 (2022)

\bibitem{laion}
Schuhmann, C., Beaumont, R., Vencu, R., Gordon, C., Wightman, R., Cherti, M., Coombes, T., Katta, A., Mullis, C., Wortsman, M., et~al.: Laion-5b: An open large-scale dataset for training next generation image-text models. Advances in Neural Information Processing Systems  \textbf{35},  25278--25294 (2022)

\bibitem{instantbooth}
Shi, J., Xiong, W., Lin, Z., Jung, H.J.: Instantbooth: Personalized text-to-image generation without test-time finetuning. arXiv preprint arXiv:2304.03411  (2023)

\bibitem{DDIM}
Song, J., Meng, C., Ermon, S.: Denoising diffusion implicit models. arXiv preprint arXiv:2010.02502  (2020)

\bibitem{vqvae}
Van Den~Oord, A., Vinyals, O., et~al.: Neural discrete representation learning. Advances in neural information processing systems  \textbf{30} (2017)

\bibitem{ELITE}
Wei, Y., Zhang, Y., Ji, Z., Bai, J., Zhang, L., Zuo, W.: Elite: Encoding visual concepts into textual embeddings for customized text-to-image generation. arXiv preprint arXiv:2302.13848  (2023)

\bibitem{attngan}
Xu, T., Zhang, P., Huang, Q., Zhang, H., Gan, Z., Huang, X., He, X.: Attngan: Fine-grained text to image generation with attentional generative adversarial networks. In: Proceedings of the IEEE conference on computer vision and pattern recognition. pp. 1316--1324 (2018)

\bibitem{Seecoder}
Xu, X., Guo, J., Wang, Z., Huang, G., Essa, I., Shi, H.: Prompt-free diffusion: Taking" text" out of text-to-image diffusion models. arXiv preprint arXiv:2305.16223  (2023)

\bibitem{RAPHAEL}
Xue, Z., Song, G., Guo, Q., Liu, B., Zong, Z., Liu, Y., Luo, P.: Raphael: Text-to-image generation via large mixture of diffusion paths. arXiv preprint arXiv:2305.18295  (2023)

\bibitem{detclip}
Yao, L., Han, J., Wen, Y., Liang, X., Xu, D., Zhang, W., Li, Z., Xu, C., Xu, H.: Detclip: Dictionary-enriched visual-concept paralleled pre-training for open-world detection. Advances in Neural Information Processing Systems  \textbf{35},  9125--9138 (2022)

\bibitem{ip-adapter}
Ye, H., Zhang, J., Liu, S., Han, X., Yang, W.: Ip-adapter: Text compatible image prompt adapter for text-to-image diffusion models. arXiv preprint arXiv:2308.06721  (2023)

\bibitem{Parti}
Yu, J., Xu, Y., Koh, J.Y., Luong, T., Baid, G., Wang, Z., Vasudevan, V., Ku, A., Yang, Y., Ayan, B.K., et~al.: Scaling autoregressive models for content-rich text-to-image generation. arXiv preprint arXiv:2206.10789  (2022)

\bibitem{GLIPv2}
Zhang, H., Zhang, P., Hu, X., Chen, Y.C., Li, L., Dai, X., Wang, L., Yuan, L., Hwang, J.N., Gao, J.: Glipv2: Unifying localization and vision-language understanding. Advances in Neural Information Processing Systems  \textbf{35},  36067--36080 (2022)

\bibitem{controlnet}
Zhang, L., Rao, A., Agrawala, M.: Adding conditional control to text-to-image diffusion models. In: Proceedings of the IEEE/CVF International Conference on Computer Vision. pp. 3836--3847 (2023)

\bibitem{uni-controlnet}
Zhao, S., Chen, D., Chen, Y.C., Bao, J., Hao, S., Yuan, L., Wong, K.Y.K.: Uni-controlnet: All-in-one control to text-to-image diffusion models. arXiv preprint arXiv:2305.16322  (2023)

\bibitem{Regionclip}
Zhong, Y., Yang, J., Zhang, P., Li, C., Codella, N., Li, L.H., Zhou, L., Dai, X., Yuan, L., Li, Y., et~al.: Regionclip: Region-based language-image pretraining. In: Proceedings of the IEEE/CVF Conference on Computer Vision and Pattern Recognition. pp. 16793--16803 (2022)

\end{thebibliography}

\end{sloppypar}

\newpage
\appendix
\title{Supplementary materials}

\author{Shengyuan Liu\inst{1,2} \and
Bo Wang \inst{3} \and
Ye Ma \inst{3} \and Te Yang \inst{1,2} \and Xipeng Cao \inst{3} \and Quan Chen \inst{3} \and Han Li \inst{3} \and Di Dong\inst{1,2} \and Peng Jiang \inst{1,2} \inst{3}}

\authorrunning{Shengyuan Liu, Bo Wang et al.}

\institute{University of Chinese Academy of Sciences, Beijing, China \and Institute of Automation, Chinese Academy of Sciences, Beijing, China \and
Kuaishou Technology, Beijing, China
}

\maketitle

\setcounter{page}{1}
\setcounter{figure}{9} 
\setcounter{table}{3} 

\begin{sloppypar}

\noindent The supplementary material contains the following parts:

\textbf{Implementation Details.} We provide more details of the experimental settings, including the network architecture, dataset, text prompt, evaluation metrics, and baseline settings. 

\textbf{Experiments.} We present additional experiments conducted on our method. We showcase the attention mechanism employed in our method and perform ablation studies on the diffusion backbone and the layers of forward guidance.

\textbf{More Examples.} We provide more examples of our method, including utilizing multi-view image inputs for single concept generation and presenting comparison results of compositional concepts generation. Furthermore, we verify the effect of fine-tuning on our proposed model.

\section{Implementation Details}
\label{sec:appendixa}
\hspace{0.75pc}\textbf{Network structure.} In our experiments, we employ Stable Diffusion v1.5 (SD v1.5) \cite{stablediffusion} as our based latent diffusion model. This framework can also be used on custom models fine-tuned from SD v1.5. We utilize OpenCLIP ViT-H/14 \cite{Openclip} as the image encoder, while incorporating IP-Adapter \cite{ip-adapter} as an additional adapter to enable image prompt capability for pretrained text-to-image diffusion models. The parameters of the IP-Adapter amount to approximately 22M. The dimension of the image features in the cross-attention blocks is 768, consistent with the dimensions of the text features in the pre-trained diffusion model.

\textbf{Dataset.} We conduct our experiments using the CustomConcept101 dataset \cite{customdiffusion}, which includes 101 concepts from various categories. More details about the dataset can be found on the \href{https://github.com/adobe-research/custom-diffusion/tree/main/customconcept101}{website}. It is important to note that the dataset includes a placeholder before each subject in the original text prompts for subject-driven generations. For example, the prompt "photo of the $<$new1$>$ person with $<$new2$>$ cat" includes the modifier tokens $<$new1$>$ and $<$new2$>$ to represent the subject attributes. In our experiments, we removed these placeholders to achieve zero-shot generation. Our goal was to associate the subject attributes with the corresponding text during the inference process using our proposed SE-Guidance.

\textbf{Text prompt.} In our experiments, we followed the settings of CustomDiffusion\cite{customdiffusion}, where the position of the subject in a sentence is predefined in CustomConcept101. In addition, we have observed that mainstream fine-tune-based methods, like Dreambooth and TextualInversion, also require predefining the location of the main subject before inserting a placeholder. When conducting large-scale inference, we consider using large language models such as ChatGPT to understand the meaning of sentences and provide the subject's position. When a word is tokenized into multiple tokens, we extract the attention map for each token that represents the subject. We then average all the attention maps to identify the key areas related to the subject.

\textbf{Evaluation metric.} In our study, we utilize CLIP-ViT-B/32 \cite{CLIP} and DINO-vits16 \cite{DINO} to compute the CLIP/DINO scores. For the initial step of the GroundingScore calculation, we employ a grounding model to establish a visual connection between natural language descriptions and the corresponding regions or objects in an image. Given an $($image, text$)$ pair, the grounding model outputs multiple sets of object boxes and noun phrases. We employ Grounding-DINO-T \cite{groundingdino} with a box threshold of 0.35 and a text threshold of 0.25. Once the target boxes are obtained, we proceed to crop the corresponding regions from the original image and compute the CLIP/DINO score, which serves as the final GroundingScore.

\textbf{Baseline model.} We follow the settings in \cite{customdiffusion} to fine-tune the baseline models including DreamBooth \cite{Dreambooth}, Textual Inversion \cite{TextualInversion}, and Custom Diffusion \cite{customdiffusion} on 2 NVIDIA V100. We trained Dreambooth with the recommended learning rate of $5\times10^{-6}$ and batchsize of 2. The training iteration is 1000 for single-subject and 2000 for multi-subject. For Textual Inversion, our experiments were conducted with a learning rate of $5\times10^{-4}$ and batchsize of 4. The number of training iterations is 1500. We trained Custom Diffusion with a batchsize of 8 and a learning rate of  $1\times10^{-5}$. The max training iterations is 500. We utilize BLIP-Diffusion \cite{Blipdiffusion} for zero-shot inference without fine-tuning. We also employ classifier-free guidance with a guidance scale of 7.5. The experiments are conducted using the Diffusers library. 

\section{Experiments}
\label{sec:appendixb}

\begin{figure*}[htb]
  \centering
   \includegraphics[width=\linewidth]{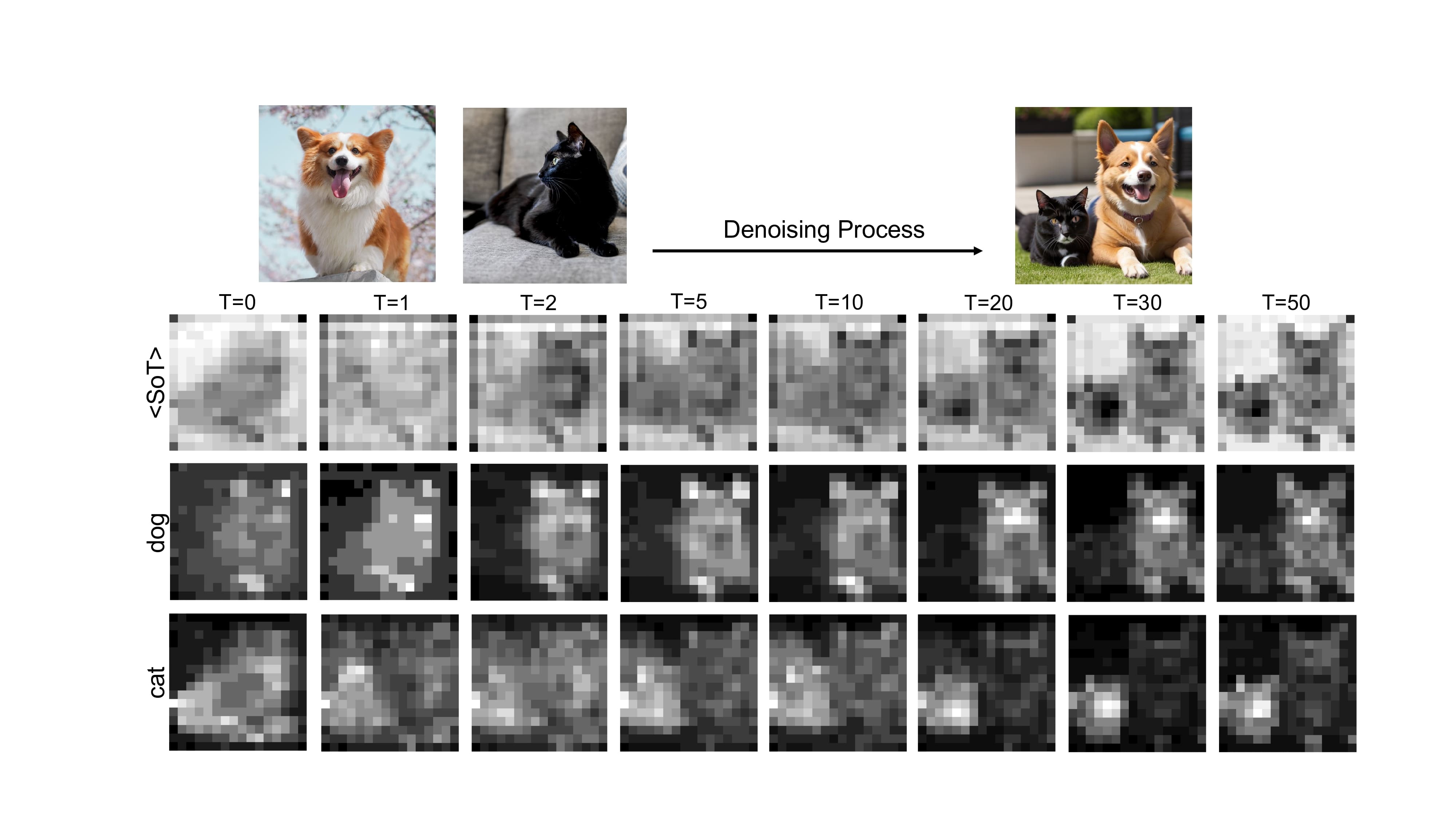}
   \caption{\textbf{Attention Maps during denoising process.} The attention maps are extracted from the cross-attention blocks of SD. The textual prompt is "\textit{A dog and a cat}". $<$SoT$>$ is the start of token in the textual embeddings. The noise scheduler is DDIM \cite{DDIM} and the number of timesteps is 50.} 
   \label{fig:attention-map}
\end{figure*}

\hspace{0.75pc}\textbf{Attention mechanism.} The core insight of our proposed method is that the attention map corresponding to the text prompt should have high attention values in specific regions of the generated image. To validate this idea, we extract and analyze the attention map related to the text prompt during our experiments. Throughout the denoising steps with backward guidance, we observe a gradual convergence of the attention map towards the corresponding locations in the synthesized image, accurately distinguishing between the dog and cat as shown in \cref{fig:attention-map}. This convergence provides precise positional information, enabling the injection of image features into the forward guidance phase, and effectively addressing the issue of attribute mixing.

Moreover, We find that the start token $<$SoT$>$ in the text feature maps carries significant semantic and layout information, particularly emphasizing the foreground details, typically represented by the union of subjects in the text prompt. Consequently, we exclude the attention map of $<$SoT$>$ and strategically re-weight the attention values to enhance the relevance of the actual prompt tokens while minimizing interference with the forward guidance.

\begin{figure*}[htb]
  \centering
   \includegraphics[width=0.75\linewidth]{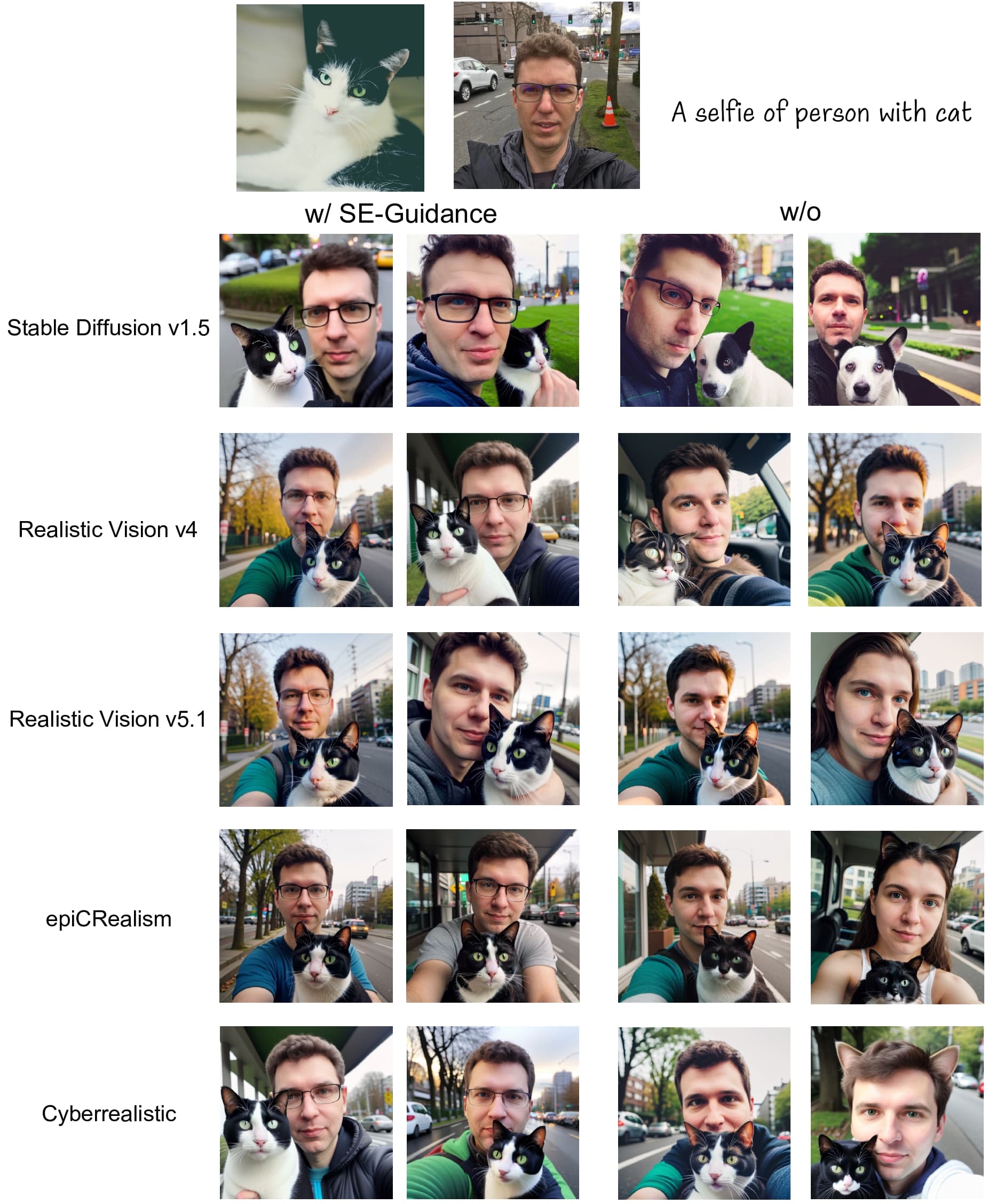}
   \caption{\textbf{Different Stable Diffusion checkpoints.} The comparison results of employing our proposed SE-Guidance on different diffusion models, fine-tuned from Stable Diffusion v1.5.} 
   \label{fig:backbone}
\end{figure*}

\textbf{Diffusion backbone.} To validate the generalizability of our proposed method, we conducted experiments using several popular realistic style models from the HuggingFace library. These models include Realistic Vision v4.0, Realistic Vision v5.1, epiCRealism, and Cyberrealistic, all of which are fine-tuned from SD v1.5 checkpoints. From the generated images, we observe that incorporating SE-Guidance into the generation process leads to the better generation of subject details, thereby effectively improving the fidelity of the subject, as shown in \cref{fig:backbone}. 

As outlined in \cref{sec:limitations}, it is important to note that while our method has resulted in improved effectiveness across all checkpoints without requiring model fine-tuning, the ultimate limitation of the model generation effect is determined by the underlying model itself. Consequently, there may still be variations in the nuances of the generated output among different checkpoints.

\textbf{Layers of forward guidance.} Our proposed forward guidance technique can be applied to different layers of the network. We conduct ablation studies to verify the effects of injecting image features at different layers. The results are shown in \cref{tab:forward}. From the results, we discover that the downsample layers play a crucial role in feature injection by focusing on the regions corresponding to the attention maps,  while the upsample layers primarily enhance details and background information. Applying forward guidance to the downsample layers effectively improves the text-image alignment in the generated images. Furthermore, when forward guidance is simultaneously applied to all layers, it achieves the best trade-off between textual semantic and subject fidelity.

\begin{table}[htb]
\caption{\textbf{Ablation study for forward guidance.} The downsample layers concentrate on the attention map regions during feature injection, while the upsample layers primarily refine details and background information.}
\label{tab:forward}
\centering
\begin{tabular}{c c c c c c}
    \toprule
Down  & Mid & Up    & CLIP-T & CLIP-GS & DINO-GS \\ 
\midrule
   \checkmark &  &                   &   0.7147     & 0.6289  &0.3570       \\
              & \checkmark &         &   0.7039      &   0.6280  & 0.3641 \\
          &           &   \checkmark &   0.7023   &   0.6147     &   0.3521     \\
    \checkmark    &   \checkmark   & &   0.7557    &  \textbf{0.6441} & \textbf{0.3731}\\
          &    \checkmark  &   \checkmark &   0.7151   &   0.6283  &  0.3570   \\
\checkmark & \checkmark & \checkmark &  \textbf{0.7714}  &   0.6391   &  0.3632   \\ 
\bottomrule
\end{tabular}

\end{table}

\section{More Examples}
\label{sec:appendixc}
\begin{figure*}[htb]
  \centering
   \includegraphics[width=0.75\linewidth]{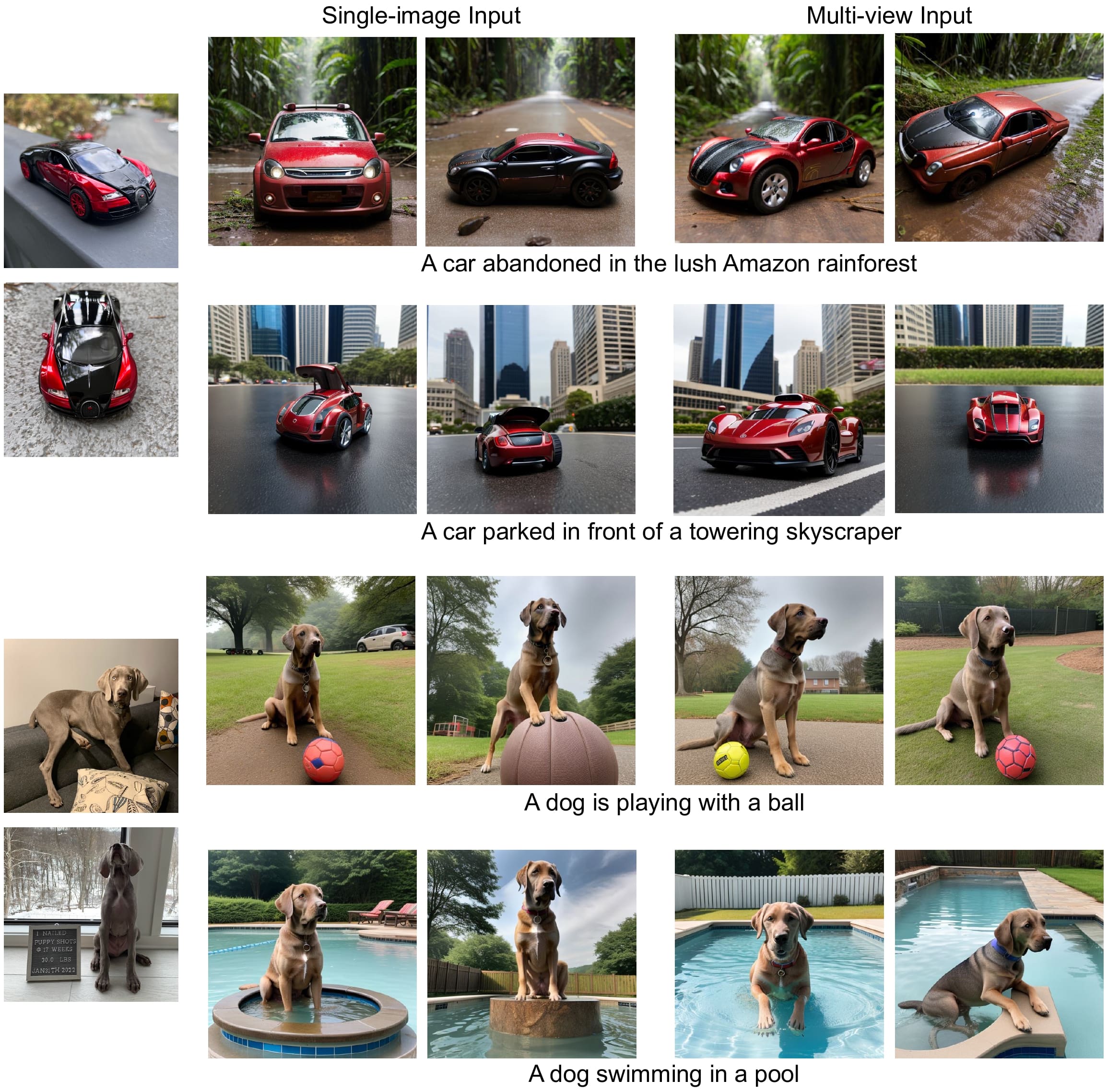}
   \caption{\textbf{Single concept generation.} Comparison of single-image and multiple-image inputs in the single-subject customization generation task.} 
   \label{fig:single-multi}
\end{figure*}

\hspace{0.75pc}\textbf{Single concept generation.} Our proposed SE-Guidance method can simultaneously support single-image and multiple-image inputs in the single-subject customization generation task. Specifically, when multiple images are used as input, we utilize the same attention map extracted from the corresponding position of the subject for forward guidance. The purpose is to provide information from different perspectives to generate better details, as shown in \cref{fig:single-multi}. Furthermore, the evaluation based on quantitative metrics, as shown in \cref{tab:compare}, indicates a slight improvement in both text-image alignment and subject fidelity.

\begin{figure*}[htb]
  \centering
   \includegraphics[width=\linewidth]{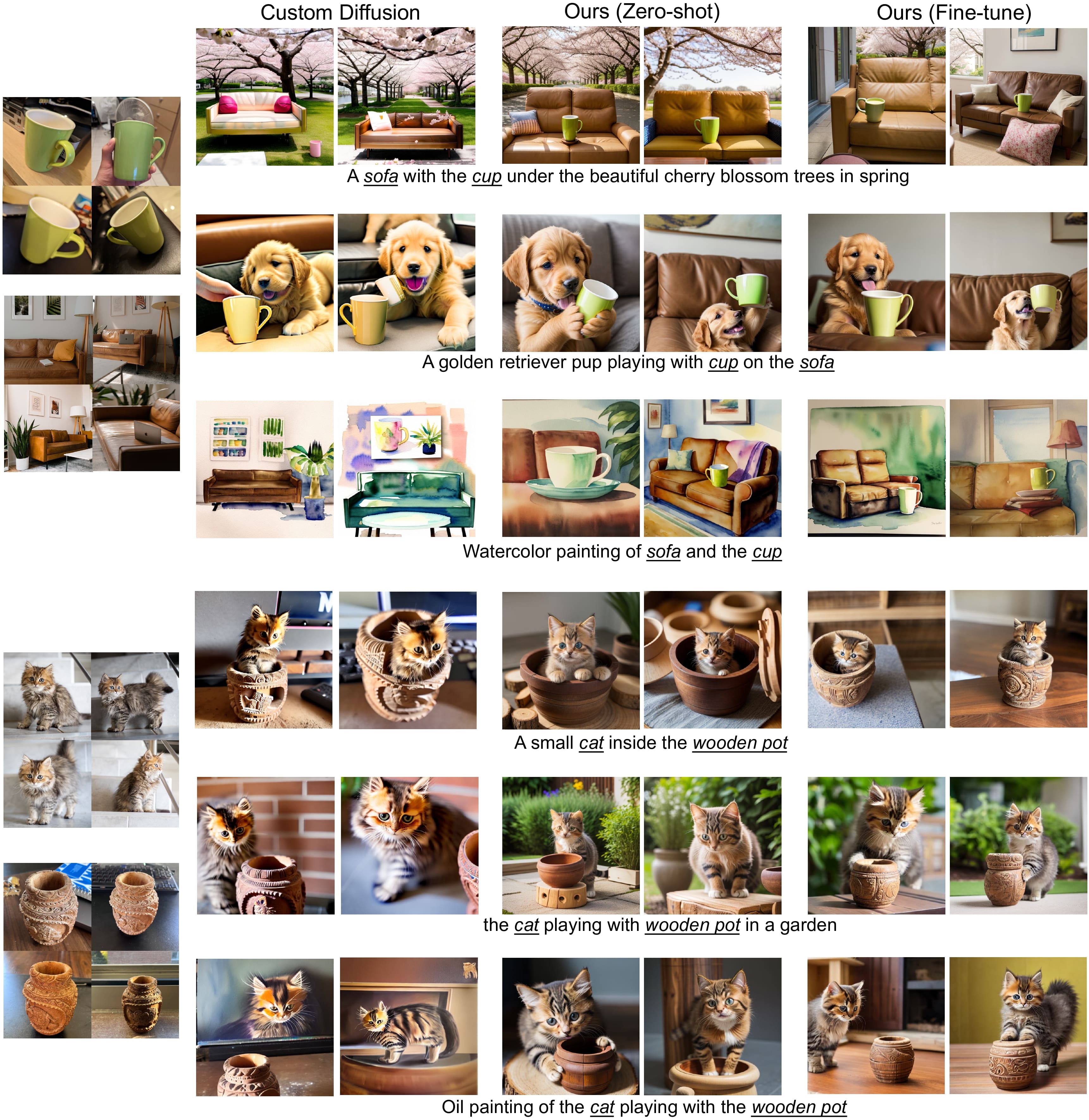}
   \caption{\textbf{Compositional concepts generation.} Comparison results between Custom Diffusion and our approaches, including zero-shot and fine-tuning strategies.} 
   \label{fig:fintune}
\end{figure*}

\textbf{Compositional concepts generation.} Our proposed method demonstrates outstanding performance by producing superior results compared to fine-tune-based methods that utilize only one image for each object. Moreover, our method significantly reduces the generation time by more than $\times 10$. However, as mentioned in \cref{sec:limitations}, our zero-shot approach is constrained by the expressive power of the generative model. While the pretrained diffusion model yields high-quality generation for common concepts such as animals, it faces considerable difficulty in zero-shot generating the correct texture for objects with unique characteristics, such as the wooden pot shown in \cref{fig:fintune}. Hence, we analyzed to examine the impact of fine-tuning on the model's generation performance.

During the fine-tuning process, we only train the parameters of the image adapter while keeping the diffusion model and image encoder frozen. We use a batch size of 8 and a learning rate of $10^{-5}$ for 1000 steps. The process takes approximately 10 minutes, which is comparable with Custom Diffusion. As shown in \cref{fig:fintune}, the experimental results demonstrate that fine-tuning effectively improves the level of detail in generated images, especially for less common objects, resulting in a noticeable enhancement compared to zero-shot generation. 

Compared to the Custom Diffusion, our approach also fine-tunes the parameters of $W_k$ and $W_v$ in the cross-attention blocks. However, we fine-tune the image branch, while Custom Diffusion fine-tune the text branch in the diffusion model. Both approaches have a similar number of fine-tuned parameters and an approximate fine-tuning time of about 10 minutes. Additionally, considering that our model does not train the backbone diffusion model, we observe that our model exhibits better text control ability. The strength of injecting image information can be controlled by hyper-parameters, reducing the risk of fine-tuning overfitting and showcasing improved versatility.

Our main contribution is to propose a novel subject-enhanced feature injection strategy, which aims to address the issues of missing objects and mixing attributes. This method proves to be effective in the inference process, regardless of whether the backbone model has undergone fine-tuning or not. Although fine-tuning is advantageous for generating details in the case of rare or highly textured objects, our proposed model plays a crucial role in decoupling attributes for compositional generation.

\end{sloppypar}

\end{document}